\documentclass[lettersize,journal]{IEEEtran}
\usepackage{amsmath,amsfonts}
\usepackage{algorithmic}
\usepackage{amssymb}
\usepackage{algorithm}
\usepackage{array}
\usepackage[caption=false,font=footnotesize,labelfont=rm,textfont=rm]{subfig}
\usepackage{textcomp}
\usepackage{stfloats}
\usepackage{url}
\usepackage{verbatim}
\usepackage{graphicx}
\usepackage{cite}
\usepackage[mathscr]{eucal}
\usepackage{bm}
\usepackage{multirow}
\usepackage[normalem]{ulem}
\useunder{\uline}{\ul}{}
\usepackage{booktabs}
\usepackage{pifont}
\usepackage{subfig}
\usepackage{threeparttable}
\usepackage{orcidlink}
\hypersetup{hidelinks}

\begin{document}

% \title{A Sample Article Using IEEEtran.cls\\ for IEEE Journals and Transactions}

\title{Learning Prompt-Enhanced Context Features for Weakly-Supervised Video Anomaly Detection}


% \author{Yujiang Pu, Xiaoyu Wu, Shengjin Wang,~\IEEEmembership{Senior Member,~IEEE}
\author{Yujiang Pu$^{\orcidlink{0009-0000-2761-5943}}$, Xiaoyu Wu$^{\orcidlink{0000-0003-3481-7820}}$, Lulu Yang, and Shengjin Wang$^{\orcidlink{0000-0001-7809-1932}}$,~\IEEEmembership{Senior Member,~IEEE}

\thanks{\emph{Corresponding author: Xiaoyu Wu.}}% <-this % stops a space
\thanks{Yujiang Pu, Xiaoyu Wu and Lulu Yang are with the School of Information and Communication Engineering, Communication University of China, Beijing 100024, China (e-mail: pyj2020@cuc.edu.cn; wuxiaoyu@cuc.edu.cn; yangll@cuc.edu.cn).}
\thanks{Shengjin Wang is with the Department of Electrical Engineering, Tsinghua University, Beijing 100084, China (e-mail: wgsgj@tsinghua.edu.cn).}}

\markboth{Journal of \LaTeX\ Class Files,~Vol.~14, No.~8, August~2021}%
{Shell \MakeLowercase{\textit{et al.}}: A Sample Article Using IEEEtran.cls for IEEE Journals}


\maketitle

\begin{abstract}
Video anomaly detection under weak supervision presents significant challenges, particularly due to the lack of frame-level annotations during training. While prior research has utilized graph convolution networks and self-attention mechanisms alongside multiple instance learning (MIL)-based classification loss to model temporal relations and learn discriminative features, these methods often employ multi-branch architectures to capture local and global dependencies separately, resulting in increased parameters and computational costs. Moreover, the coarse-grained interclass separability provided by the binary constraint of MIL-based loss neglects the fine-grained discriminability within anomalous classes. In response, this paper introduces a weakly supervised anomaly detection framework that focuses on efficient context modeling and enhanced semantic discriminability. We present a Temporal Context Aggregation (TCA) module that captures comprehensive contextual information by reusing the similarity matrix and implementing adaptive fusion. Additionally, we propose a Prompt-Enhanced Learning (PEL) module that integrates semantic priors using knowledge-based prompts to boost the discriminative capacity of context features while ensuring separability between anomaly sub-classes. Extensive experiments validate the effectiveness of our method's components, demonstrating competitive performance with reduced parameters and computational effort on three challenging benchmarks: UCF-Crime, XD-Violence, and ShanghaiTech datasets. Notably, our approach significantly improves the detection accuracy of certain anomaly sub-classes, underscoring its practical value and efficacy. Our code is available at: \url{https://github.com/yujiangpu20/PEL4VAD}.
\end{abstract}

% To further enhance performance, we introduce a Score Smoothing (SS) module in the testing phase, designed to suppress individual bias and minimize false alarms. 


\section{Introduction}

\IEEEPARstart{V}{ideo} Anomaly Detection (VAD) is the process of identifying unusual events or behaviors that diverge from normal patterns in video streams \cite{Adam2008, ramachandra2020survey, liu2023generalized}. With the increasing ubiquity of surveillance cameras, relying solely on human supervision is inadequate for meeting the demands of practical applications. Consequently, there's a burgeoning need for efficient and accurate automated VAD methods. These methods are crucial across various sectors, including security, industrial monitoring, healthcare, and social media analysis, where they help mitigate potential threats and enhance operational efficiency, thereby holding significant practical value.

Unlike classical action recognition tasks, gathering relevant videos for anomaly detection is challenging due to the rarity and sensitive nature of anomalous events. The prevailing approach treats anomaly detection as a semi-supervised learning task \cite{Pang2021}, where the model learns the patterns of normal behavior during training and identifies deviations as anomalies \cite{Hasan2016, Gong2019, Park2020, Luo2021, Chen2022, Wang2023}. These approaches don't require detailed annotations but often mistakenly identifies unseen samples as anomalies due to the incomplete representation of normal patterns, resulting in high false alarm rates.

Recently, there's been a notable shift toward developing more robust algorithms for anomaly detection, with a specific focus on weakly supervised methods \cite{Sultani2018, Wu2021, Tian2021, wu2022self, zhang2023exploiting}. These offer several advantages over semi-supervised approaches: 1) They incorporate both normal and abnormal videos in training, enabling more discriminative representation learning. 2) They require only video-level annotations, indicating the presence but not the exact timing of anomalies. 3) They facilitate the creation of large-scale datasets due to less stringent annotation requirements. Owing to these benefits, weakly supervised methods have demonstrated superior performance, significantly outshining their semi-supervised counterparts.

To effectively model the temporal dynamics of anomalous events, which vary significantly in duration, it's crucial to understand the contextual information within video snippets. Existing methods, such as graph convolutional networks \cite{defferrard2016convolutional} and self-attention mechanisms \cite{vaswani2017attention}, aim to capture these temporal relationships. However, each has its limitations: graph convolutions only consider local node aggregates, potentially missing the broader context of anomalies, while self-attention mechanisms, despite capturing global correlations, might introduce irrelevant noise. To refine this understanding, Wu \emph{et al}. \cite{Wu2020} and Tian \emph{et al}. \cite{Tian2021} devised networks that integrate local and global contexts through parallel structures. Although these approaches enhance context modeling by mitigating long-range noise interference, they come at the cost of increased computational complexity and parameter count, which can lead to overfitting and hinder practical deployment.

% \begin{figure}[!t]
% \centering
% \subfloat[]{\includegraphics[width=0.9\linewidth]{1-1.png}}
% \hfil
% \subfloat[]{\includegraphics[width=0.9\linewidth]{1-2.png}}
% \caption{Complex abnormal events from different acquisition sources. (a) UCF-Crime \cite{Sultani2018}. (b) XD-Violence \cite{Wu2020}.}
% \label{fig_sim}
% \end{figure}

\begin{figure}[!t]
\centering
\subfloat[]{\includegraphics[width=0.44\linewidth]{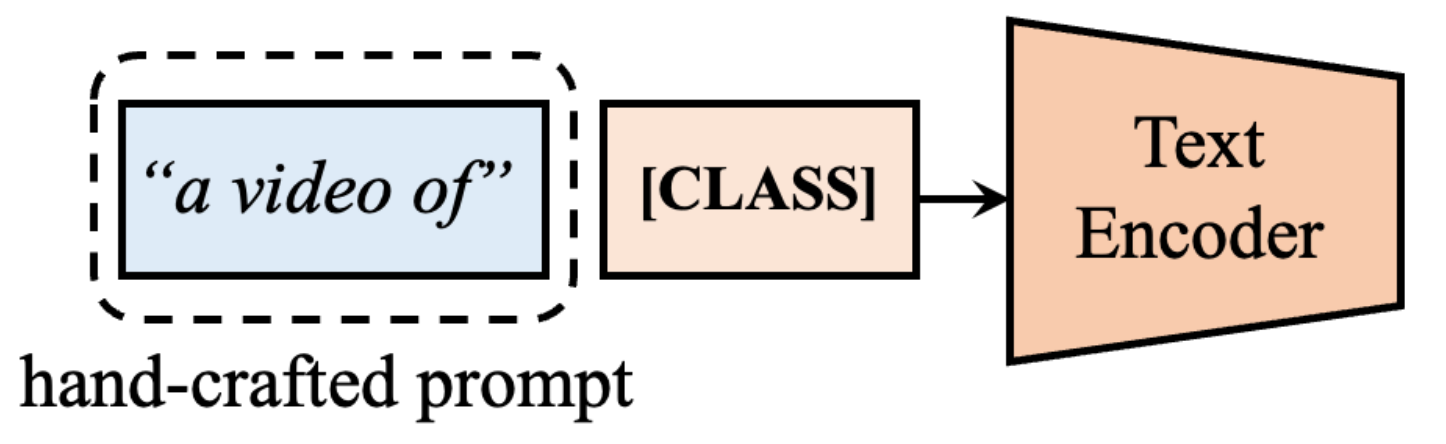}%
\label{fig_first_case}}
\hfil
\subfloat[]{\includegraphics[width=0.46\linewidth]{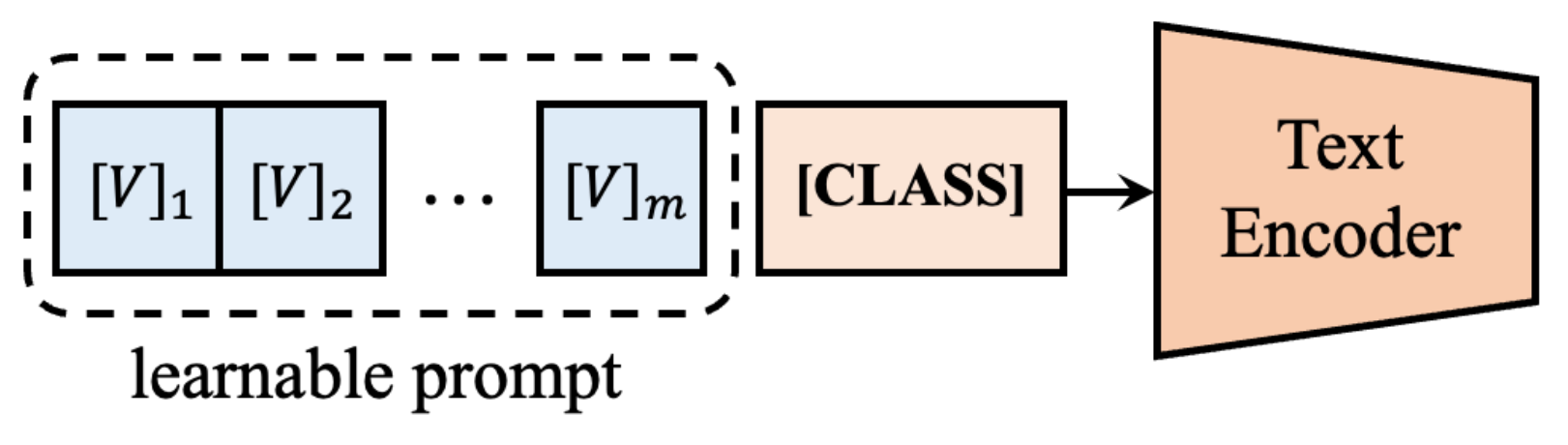}%
\label{fig_second_case}}
\\ 
\subfloat[]{\includegraphics[width=\linewidth]{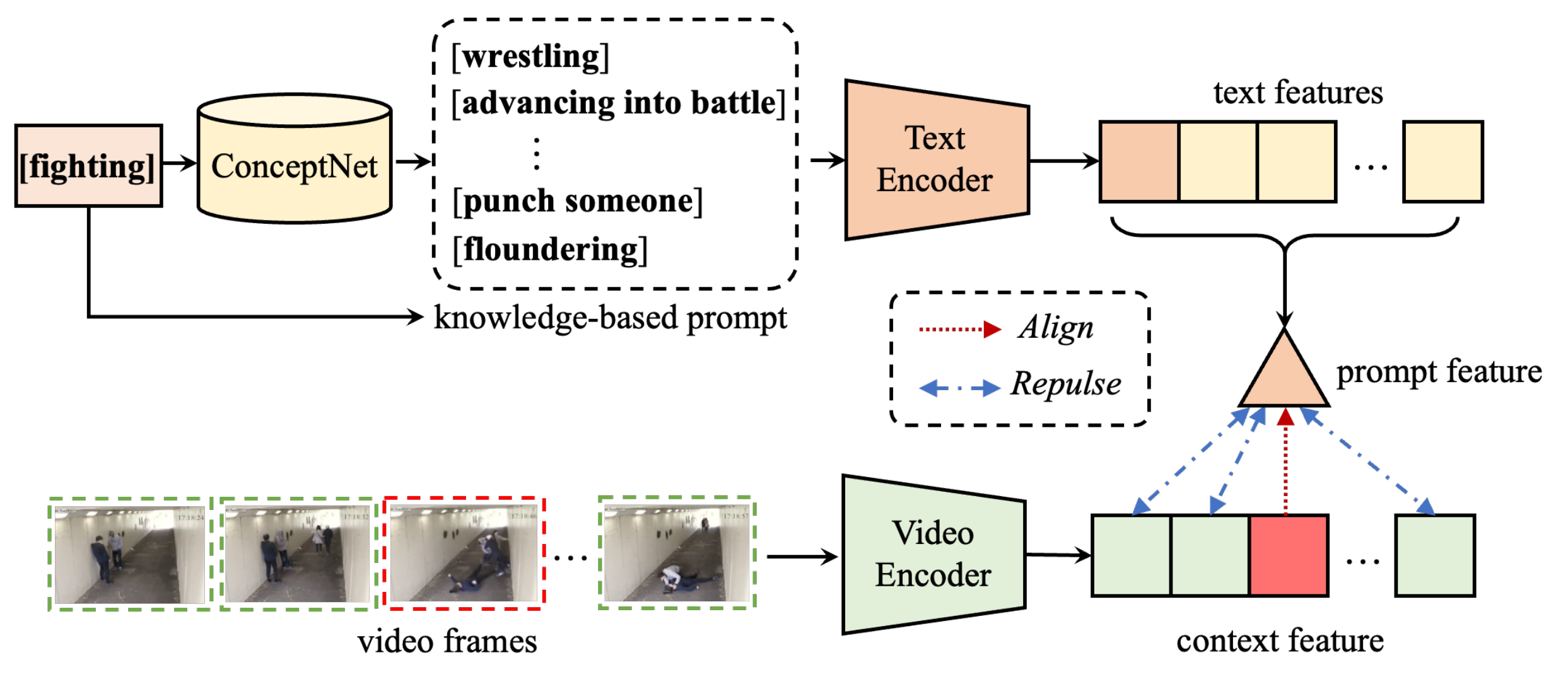}%
\label{fig_third_case}}
\caption{Comparison of different methods for prompt construction. (a) Hand-crafted prompt. (b) Learnable prompt. (c) Knowledge-based prompt (Ours). The context features are enhanced by cross-modal alignment with the corresponding prompt features.}
\label{fig_sim}
\end{figure}

Furthermore, the inherent diversity and complexity of anomalies pose additional challenges. Anomaly datasets often showcase large intra-class variance and minimal inter-class variation, with examples ranging from fixed-camera surveillance footage to artistically varied film and TV content. Addressing this requires a robust detection pipeline capable of navigating these complexities. While recent methods aim to construct a compact normal manifold \cite{Wu2021} and introduce additional binary constraints \cite{Park2023} for discriminative learning, they often neglect the distinct visual characteristics within anomaly classes and fail to capture the specific semantics of various anomalies. This oversight leads to poor interpretability and an increased risk of misclassification.

In light of these limitations, our motivation is twofold: 1) to develop an approach that balances the need for detailed temporal modeling with computational efficiency, thereby addressing the high parameter count and low efficiency issues inherent in multi-branch structures. 2) to construct a method that recognizes and interprets the diversity and complexity of anomaly semantics more effectively than existing models.

In this paper, we firstly introduce a Temporal Context Aggregation (TCA) module designed to capture temporal relations across video snippets more efficiently than the current parallel architectures \cite{Tian2021} \cite{Wu2020} \cite{Zhong2019} \cite{zhou2023dual}. This module distinguishes itself by reutilizing the similarity matrix, thereby reducing the computational load and parameter count. It also incorporates a learnable factor to adaptively fuse local and global contexts and introduces a Dynamic Position Encoding (DPE) with a parametric kernel function to accurately model the relative positions of video snippets. As evidenced by the results in Section \uppercase\expandafter{\romannumeral4}, our TCA module demonstrates superior performance with a more streamlined and cost-effective design compared to existing methods. 

Secondly, we propose a Prompt-Enhanced Learning (PEL) module, drawing inspiration from prompt learning \cite{Liu2023}. Unlike existing methods \cite{radford2021learning, wang2021actionclip, Zhou2022, ju2022prompting} that rely on hand-crafted or learnable prompts, our approach leverages an external knowledge base, i.e., ConceptNet \cite{Speer2017}, to construct prompt templates, as demonstrated in Figure 1. These templates are contextually rich and possess a degree of interpretability, offering a nuanced understanding of the specific semantics of anomalies. By aligning anomalous contexts with corresponding prompt features and distancing non-anomalous contexts, our PEL module injects rich semantics into visual features, enhancing fine-grained discriminability and inter-class separability. This innovative approach fosters the development of a more discerning decision boundary in the embedding space.

Finally, considering that anomalies in videos generally extend across several frames, we utilize a plug-and-play score smoothing (SS) strategy during the testing phase to address potential mis-classifications due to transient disturbances such as frame jitter or switching. This module applies average pooling to consecutive anomaly scores, thereby suppressing individual biases and mitigating the impact of ephemeral noise inconsistent with the typical duration of an anomaly. 

% Score smoothing enhances the model's detection capabilities in novel scenarios, thereby improving overall performance without causing overfitting to the training data.

The main contributions of this paper are succinctly outlined as follows:

\begin{itemize}
  \item [1)]
  We introduce a Temporal Context Aggregation (TCA) module that leverages a reused similarity matrix and adaptive fusion for efficient and effective video anomaly detection. This module demonstrates enhanced encoding performance while significantly reducing the parameter count and computational expense.
  \item [2)]
  We present a Prompt-Enhanced Learning (PEL) module designed to augment fine-grained contextual understanding. This is achieved through the construction of knowledge-based prompts, precise context separation, and cross-modal alignment, thereby enriching the semantic discrimination of the model.
  \item [3)]
  Our model exhibits competitive performance across three benchmark datasets: UCF-Crime, XD-Violence, and ShanghaiTech. Notably, it shows an approximate 10\% improvement in detection accuracy for certain fine-grained anomalies, underscoring its practical efficacy.
\end{itemize}

The remainder of the paper is structured as follows: Section \uppercase\expandafter{\romannumeral2} reviews related work in video anomaly detection and prompt learning. Section \uppercase\expandafter{\romannumeral3} delineates the proposed method in detail. Section \uppercase\expandafter{\romannumeral4} presents comparative results with state-of-the-art methods on benchmark datasets and validates the effectiveness of each component through ablation studies. Finally, Section \uppercase\expandafter{\romannumeral5} concludes the paper, reflecting on findings and future directions.

\section{Related Work}
\subsection{Video Anomaly Detection}
Early studies \cite{Mehran2009, Zhao2011, Lu2013, WeixinLi2014} treated anomaly detection as a semi-supervised task, utilizing statistical models with hand-crafted features. These approaches, however, suffered from limited representational capacity, leading to suboptimal robustness and generalization. Recent advancements in deep learning have spurred the development of more sophisticated anomaly detection techniques. Some methods \cite{Xu2017, Sabokrou2018, Wu2019, Wang2019} aim to create a one-class classifier that delineates normality in a latent space, identifying deviations as anomalies. Yet, without prior knowledge of anomalies, these models may mislabel complex or unseen normal samples as anomalous. Other strategies \cite{Liu2018, ZaighamZaheer2020, medel2016anomaly, liu2023amp} postulate that anomalies manifest as larger reconstruction or prediction errors, assuming normality can be accurately reconstructed or predicted.
However, these methods are prone to overfitting and may reconstruct or predict anomalies well.

% This approach is often implemented using deep models like U-Net \cite{Ronneberger2015}, GAN \cite{Goodfellow2020}, and ConvLSTM \cite{shi2015convolutional}, which are susceptible to overfitting and may inaccurately classify anomalies as normal.

Recent research has pivoted towards weakly supervised anomaly detection, leveraging both normal and abnormal samples with video-level annotations. These methods generally categorize into two paradigms: one-stage methods utilizing Multiple Instance Learning (MIL) and two-stage self-training strategies. Sultani \emph{et al}. \cite{Sultani2018} introduced a deep MIL model optimizing a margin ranking loss to enhance the differentiation between positive and negative snippets. Tian \emph{et al}. \cite{Tian2021} developed Robust Temporal Feature Magnitude (RTFM) learning, detecting subtle anomalies through feature similarity and temporal continuity. Cho \emph{et al}. \cite{cho2023look} devised Class-Activate Feature Learning (CLAV) to extract distinctive features and employed relative distance learning to accentuate their differences, alongside the Context-Motion Interrelation Module (CoMo) for analyzing scene-related motion anomalies. Conversely, Lv \emph{et al}. \cite{lv2023unbiased} proposed an unbiased MIL approach that considers both confident and ambiguous snippets, enhancing the detector's ability to differentiate between normal and abnormal events without relying on contextual biases.

Two-stage self-training methods have emerged to generate high-confidence pseudo-labels for video snippets, recasting weakly supervised anomaly detection as a supervised task with noisy labels. Zhong \emph{et al}. \cite{Zhong2019} employed a graph convolution network to refine these labels and iteratively enhance the anomaly classifier. Feng \emph{et al}. \cite{Feng2021} proposed utilizing multi-instance pseudo label generation to fine-tune feature encoders, thereby yielding task-specific discriminative features. Li \emph{et al}. \cite{li2022self} introduced Multi-Sequence Learning (MSL), a method that incrementally optimizes reduced sample lengths to refine localization boundaries. Zhang \emph{et al}. \cite{zhang2023exploiting} introduced a multi-head module to generate varied pseudo labels and an iterative uncertainty-based training strategy that refines the process by prioritizing clips with lower uncertainty.

In parallel, several studies have leveraged multimodal information to enhance video anomaly detection. Wu \emph{et al}. \cite{Wu2020} proposed HL-Net, integrating appearance, motion, and audio for a comprehensive multimodal approach. Yu \emph{et al}. \cite{Yu2022} utilized a dual-stream network to improve instance clustering and employ self-distillation to minimize the semantic gap between unimodal and audio-visual features. Wei \emph{et al}. \cite{wei2022msaf} introduced the MSAF framework, employing multimodal MIL ranking loss to generate pseudo clip-level labels and a supervise-attention regression loss in feature learning to foster implicit alignment between different modalities. Pu \emph{et al}. \cite{Pu2022cma} enhanced contextual video representation through cross-modal interactions between audio and visual elements, utilizing temporal convolution to compute high-confidence scores for specific events like violence.

While numerous methods have explored temporal relation modeling, many depend on parallel branches that introduce additional parameters and computational demands. In contrast, our approach presents two significant advantages. Firstly, the TCA module capitalizes on the reuse of the similarity matrix to concurrently capture local-global dependencies. This strategy outperforms methods requiring extra branches \cite{Tian2021}, \cite{Wu2020}, \cite{zhou2023dual}, and \cite{Zhen2021}, while minimizing both parameters and computational load. Secondly, diverging from techniques that construct normality prototypes from the data itself, like \cite{Wu2021} and \cite{Park2023}, our PEL module introduces anomaly priors into the model using external knowledge. This approach not only enriches the semantics of visual features but also notably advances the fine-grained anomaly detection capabilities of our model.

\subsection{Prompt Learning in Video Understanding}
Prompt learning, initially introduced in natural language processing, aims to utilize pre-trained models for few-shot or zero-shot scenarios by designing appropriate prompt templates. This technique has recently been extended to video understanding tasks. Wang et al. \cite{wang2021actionclip} demonstrated this by aligning video clips with prompt embeddings of category labels for action recognition. They incorporated category labels as prefixes, fill-in-the-blanks, and suffixes into prompt templates, thereby significantly outperforming single category prompts. Similarly, Ju et al. \cite{ju2022prompting} crafted prompt templates by merging category labels with learnable vectors, examining the impact of label positioning within the templates. Despite these advancements, manual prompt design remains laborious, and models are highly sensitive to template content \cite{gao2020making}. While parametric prompts eliminate the need for linguistic priors, they often result in abstract and cryptic strings.

Recently, Yao et al. \cite{yao2022detclip} leveraged WordNet's \cite{Miller1995} definitions to formulate prompt templates and created additional category label descriptions using a concept dictionary. This method significantly advanced open-vocabulary object detection tasks. Inspired by their work, we introduce a novel prompt-enhanced learning strategy for video anomaly detection. Unlike previous methods, we extract anomaly-relevant, class-specific concepts from ConceptNet \cite{Speer2017} to augment contextual information. Our objective is to improve visual features through external prompts. We achieve this by infusing semantic information into visual features via cross-modal alignment, rather than bidirectional matching, ensuring reliance solely on visual features during testing.

\section{The Proposed Method}
In this section, we will introduce the proposed method for weakly-supervised video anomaly detection. Firstly, we present the overall framework, followed by a detailed elaboration of the core components.

\subsection{Overall Framework}

The overall framework of our method is shown in Figure 2. Specifically, an untrimmed video is first divided into non-overlapping snippets by a sliding window of 16 frames. Subsequently, we use a pre-trained I3D network \cite{Carreira2017} to extract the snippet features $\mathbf{X}$, which are fed into the TCA module to generate context features $\mathbf{X}^c$. After that, feature reduction is achieved through a two-layer multilayer perceptron (MLP), with the PEL module applied to the middle layer to learn discriminative features $\mathbf{X}^e$ via knowledge-based prompts. Finally, the classifier predicts snippet-level anomaly scores $\mathbf{S}$. During the training phase, the MIL-based loss function converts the snippet-level scores into bag-level predictions to learn high activations for anomalous cases, where the subscripts $a$ and $n$ denote abnormal and normal videos, respectively.

\begin{figure*}[!t]
\centering
\includegraphics[width=\linewidth]{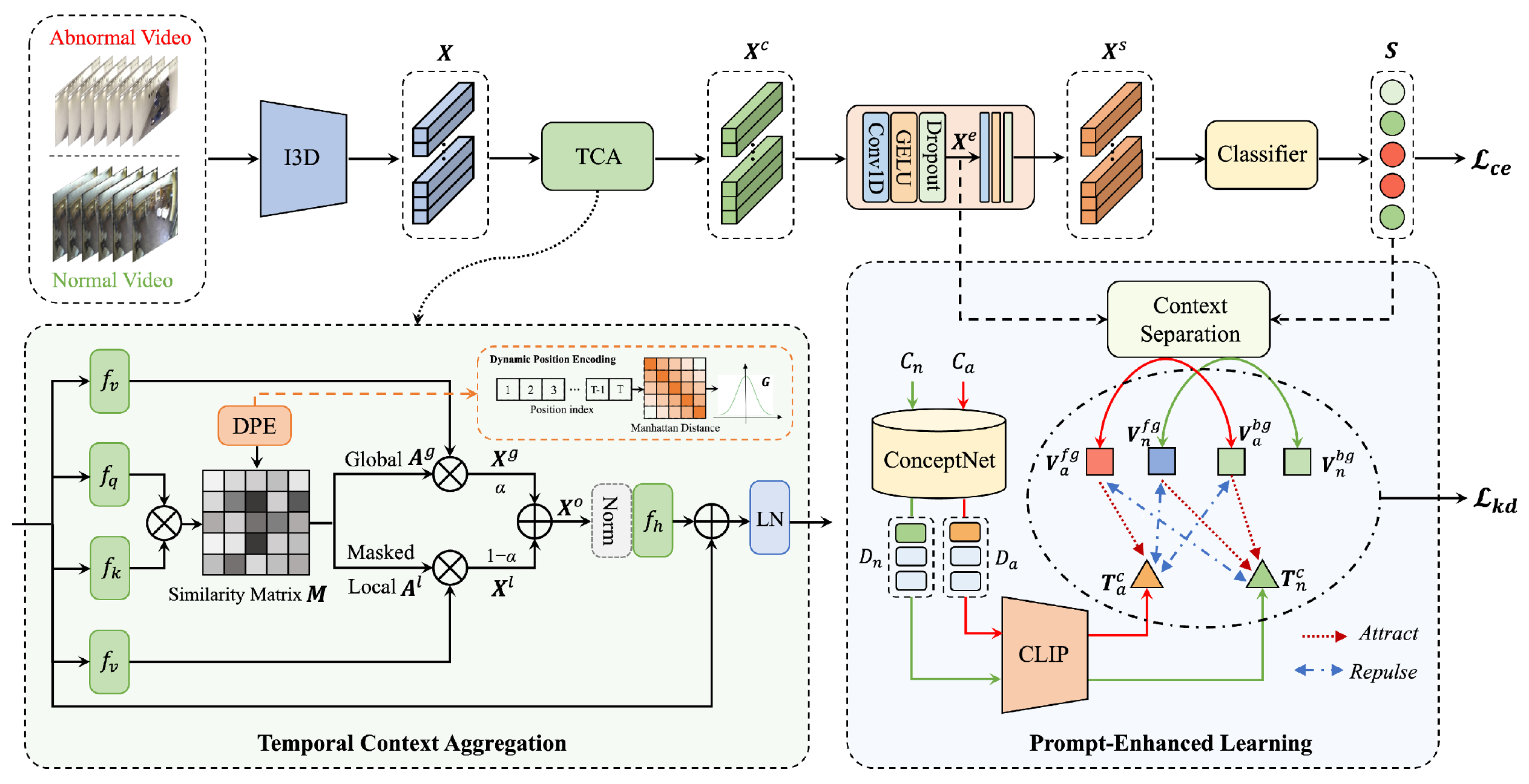}
\caption{Overview of the proposed framework. We first process untrimmed videos using a pre-trained I3D network to extract snippet features. The TCA module then simultaneously captures both local and global contexts. High-level representations are derived using a two-layer MLP, with the PEL module operating in the middle layer to learn fine-grained semantics for context features. Finally, a causal convolution layer functions as a classifier to predict snippet-level anomaly scores. During the training stage, the model is optimized using both cross entropy loss $\mathscr{L}_{ce}$ and KL-divergence loss $\mathscr{L}_{kd}$.}
\end{figure*}

% The goal of anomaly detection is to determine whether a video contains anomalies while localizing the interval. Under the weakly supervised setting, a video bag $\mathcal{V}=\{v_i\}^T_{i=1}$ and its corresponding video-level label $y\in\{0,1\}$ are given, where $v_i$ denotes non-overlap snippets and $y$ indicates the presence of anomalies in the video, respectively. The predicted anomaly score of the video is denoted as $\mathcal{S}=\{s_j\}^T_{j=1}$, where $s_j$ is the anomaly score of the $j^{th}$ snippet. 

\subsection{Temporal Context Aggregation Module}
As previously discussed, the Temporal Context Aggregation (TCA) module, depicted in Figure 2, diverges from the conventional multi-branch parallel frameworks used in existing methods. It concurrently models global-local dependencies by leveraging a similarity matrix and adaptive fusion. The snippet feature $\mathbf{X}$ is first projected to the latent space by different linear layers, and the similarity matrix $\mathbf{M}$ is obtained by computing the inner product as follows:
\begin{equation}
\mathbf{M}=f_q(\mathbf{X})\cdot f_k(\mathbf{X})^\top,
\end{equation}
\begin{equation}
\mathbf{A}^g = \text{softmax}(\frac{\mathbf{M}}{\sqrt{D_h}}),
\end{equation}
\begin{equation}
\mathbf{X}^g=\mathbf{A}^g\cdot f_v(\mathbf{X}),
\end{equation}
\noindent where $f_q(\cdot)$, $f_k(\cdot)$ and $f_v(\cdot)$ are three different linear layers, $\top$ refers to the transpose operation, and $D_h$ is the hidden dimension in the latent space. Softmax normalization is then applied to generate the global attention map $\mathbf{A}^g$. The projected snippet feature is re-weighted by the attention map to obtain the global context feature $\mathbf{X}^g$.

% \begin{figure}[!t]
% \centering
% \includegraphics[width=\linewidth]{method_fig/TCA.png}
% \caption{Structure of the TCA module. DPE denotes dynamic position encoding.}
% \end{figure}

Although the above operation facilitates global context modeling, it inevitably introduces long-range noise. To address this issue, we implement local context calibration by reusing the similarity matrix with a masking window, which can be formulated as:
\begin{equation}
\Tilde{\mathbf{M}}_{ij}=
	\begin{cases}
	\mathbf{M}_{ij}, \quad \text{if} \text{ } j\in\left[\text{max}(0,i-\left\lfloor\frac{w}{2}\right\rfloor), \text{min}(i+\left\lfloor\frac{w}{2}\right\rfloor,T) \right]  \\
	-\infty, \quad \text{otherwise}
	\end{cases}
\end{equation}
\begin{equation}
\mathbf{A}^l = \text{softmax}(\frac{\Tilde{\mathbf{M}}}{\sqrt{D_h}}),
\end{equation}
\begin{equation}
\mathbf{X}^l=\mathbf{A}^l\cdot f_v(\mathbf{X}),
\end{equation}
\noindent where $w$ is the window size of the mask and $T$ is the maximum length of the input sequence. Eq.\,(4) ensures that the $i^{th}$ snippet interacts only with its neighborhood of window size $w$. The lower bound of this window is the earliest moment that can be historically observed, and the upper bound is the maximum length of the sequence. Similarly, the projected snippet feature is re-weighted with the attention map $\mathbf{A}^l$ to obtain local calibration features $\mathbf{X}^l$, which can effectively capture slight changes and achieve feature enhancement in the local neighborhood.

% \begin{figure}[!t]
% \centering
% \includegraphics[width=0.9\linewidth]{method_fig/DPE.pdf}
% \caption{Illustration of Dynamic Position Encoding}
% \end{figure}

Subsequently, we resort to a learnable fashion rather than direct concatenation or average pooling to achieve local-global contextual adaptive fusion, allowing the model to dynamically balance the importance of global temporal patterns and local nuances. The fusion process is formulated as follows:
\begin{equation}
\mathbf{X}^o = \alpha\cdot\mathbf{X}^g+(1-\alpha)\cdot\mathbf{X}^l,
\end{equation}
\begin{equation}
\mathbf{X}^c = \text{LN}(\mathbf{X}+f_h(\text{Norm}(\mathbf{X}^o))),
\end{equation}
\noindent where $\alpha$ and $1-\alpha$ denote the global and local weight for context fusion, respectively, and $\text{Norm}(\cdot)$ denote a combination of power normalization \cite{Yu2017} and L2 normalization. Then a linear layer $f_h(\cdot)$ followed by residual connection and layer normalization $\text{LN}(\cdot)$ is applied to obtain context feature $\mathbf{X}^c$.

In addition, considering the importance of position information, we introduce Dynamic Position Encoding (DPE) to model the relative distances of snippets, formulated as:
\begin{equation}
\mathbf{G}=\text{exp}(-|{\gamma{(i-j)}^{2}}+\beta|),
\end{equation}
\noindent where $i$ and $j$ denote the absolute positions of two snippets, and $\gamma$ and $\beta$ are learnable weights and bias terms. In particular, the DPE is embedded into the similarity matrix $\mathbf{M}$ as a location prior, i.e., $\mathbf{M} \gets \mathbf{M}+\mathbf{G}$, thus avoiding affecting the original feature distribution. Unlike fixed position encodings in \cite{vaswani2017attention}, DPE adapts to varying video lengths due to its dynamic nature. Also, the Gaussian-like kernel of DPE inherently suppresses the influence of long-distance noise, which emphasizes closer snippet relationships over distant ones implies an innate sensitivity to non-linear patterns.

\subsection{Multilayer Perceptron and Classifier}
To obtain high-level semantic representations, a two-layer MLP is utilized for feature reduction. Each Conv1D layer is followed by a GELU activation and dropout operation. This process is denoted as follows:
% \begin{equation}
% \mathbf{X}^c = \mathbf{W}_2(\mathbf{W}_1\mathbf{X}^o +\mathbf{b}_1)+\mathbf{b}_2,
% \end{equation}
\begin{equation}
\begin{aligned}
\mathbf{X}^e = \text{Dropout}(\text{GELU}(\text{Conv1D}(\mathbf{X}^c))),\\
\mathbf{X}^s = \text{Dropout}(\text{GELU}(\text{Conv1D}(\mathbf{X}^e))).
\end{aligned}
\end{equation}

Finally, a causal convolution layer is employed to predict snippet-level anomaly scores, which considers both current and historical observations to obtain more reliable results. The classifier is formulated as:
% \begin{equation}
%     \mathbf{S}=\sigma ({{f}_{t}}([\mathbf{x}_{i-\Delta t}^{c},...,\mathbf{x}_{i}^{c}]))
% \end{equation}
\begin{equation}
    \mathbf{S}=\sigma ({{f}_{t}}(\mathbf{X}^s)),
\end{equation}
\noindent where $f_t(\cdot)$ is the causal convolution layer with kernel size $\Delta t$, $\sigma(\cdot)$ is the sigmoid function, and $\mathbf{s}_i$ is the anomaly score of the $i^{th}$ snippet.

Following \cite{Wu2021}, we adopt the MIL-based loss as the basic objective function. Specifically, we determine the video-level prediction $p_i$ by taking the mean value of the top-\emph{k} anomaly scores. For positive bags, we set $k=\left\lfloor {T}/{16}\;+1 \right\rfloor$, and for negative bags, we set $k=1$. Given a mini batch containing $B$ samples with video-level ground-truth $y_i$, the binary cross-entropy is formulated as:
\begin{equation}
    {\mathscr{L}_{ce}}=\sum\limits_{n=1}^{B}{-{{y}_{i}}\,\text{log} ({{p}_{i}})}.
\end{equation}

% \begin{equation}
%    M(i,j) = \sum_{k=2}^{N} \left( 1 - \frac{\lambda}{2} \right) \cdot \left| \left\| I_k(i,j) \right\|_2 - \left\| I_{k-1}(i,j) \right\|_2 \right|
% \end{equation}

% \begin{figure}[!t]
% \centering
% \includegraphics[width=\linewidth]{method_fig/MLP.png}
% \caption{Structure of MLP, CC module (left), and MIL-based loss (right).}
% \end{figure}

% \begin{figure}[!t]
% \centering
% \includegraphics[width=\linewidth]{method_fig/PEL.pdf}
% \caption{Illustration of PEL module. Knowledge-based prompts are first constructed via ConceptNet. Video-level foreground and background are then separated. Cross-modal alignment encourages video-prompt pairs with the same semantics to converge, while negative pairs with different semantics to be repulsed, further forming discriminative decision boundaries.}
% \end{figure}

\subsection{Prompt-Enhanced Learning}
The PEL module aims to enrich the visual representations by incorporating knowledge-based contextual information, thereby enhancing the model's ability to identify anomalies in intricate scenarios. Specifically, the PEL module comprises three steps, namely prompt construction, context separation, and cross-modal alignment, as depicted in Figure 2.

\subsubsection{Prompt Construction}
Considering the versatility of prompts, we first select 12 common relations\footnote{The relations include: \emph{IsA}, \emph{PartOf}, \emph{HasA}, \emph{UsedFor}, \emph{CapableOf}, \emph{Causes}, \emph{HasSubevent}, \emph{HasPrerequisite}, \emph{HasProperty}, \emph{DefinedAs}, \emph{MannerOf} and \emph{SimilarTo}. The detailed description of each relation can be found at: \href{https://github.com/commonsense/conceptnet5/wiki/Relations}{https://github.com/commonsense/conceptnet5/wiki/Relations}.} from ConceptNet as the pre-retrieval semantics, followed by choosing the top five items with the highest occurrence frequency across all categories as the retrieval relations, denoted as $\{r_j\}_{j=1}^R$. The concept dictionary $\mathcal{D}$ is then constructed by retrieving all edges of the relation $r_j$ established with a given class $c$ as the head or tail node, as illustrated in Figure 3. The non-category node in each edge is used as the key, and the relevance score of the edge is taken as the value. To remove noisy entries that may interfere with the semantics of anomalies, we first eliminate concepts with relevance scores less than or equal to 0 (Step\,1). Next, the remaining entries are filtered by using the average of the relevance scores as the threshold (Step\,2). In general, the higher the relevance score, the stronger the semantic relationship between the nodes.

After obtaining the concept dictionary, we use a pre-trained CLIP model \cite{radford2021learning} to extract the corresponding prompt representations. For a given class $c$, a set of keys $\{k_i^c\}_{i=1}^N$ from the concept dictionary is first extracted as the context prompt, which is then separately fed into the text encoder to extract 512-dimensional feature vectors  $\{\mathbf{K}_i^c\}_{i=1}^N$. Finally, the average of all feature vectors is regarded as the knowledge-based prompt feature, denoted as follows:
\begin{equation}
    \mathbf{T}^{c}=\frac{1}{N}\underset{i=1}{\overset{N}{\mathop \sum }}\,\mathbf{K}_{i}^{c},
\end{equation}
\noindent where $N$ denotes the number of keys. The prompt representation incorporates relevant concepts from multiple relations and facilitates the semantic enhancement of visual features.

\begin{figure}[!t]
\centering
\includegraphics[width=\linewidth]{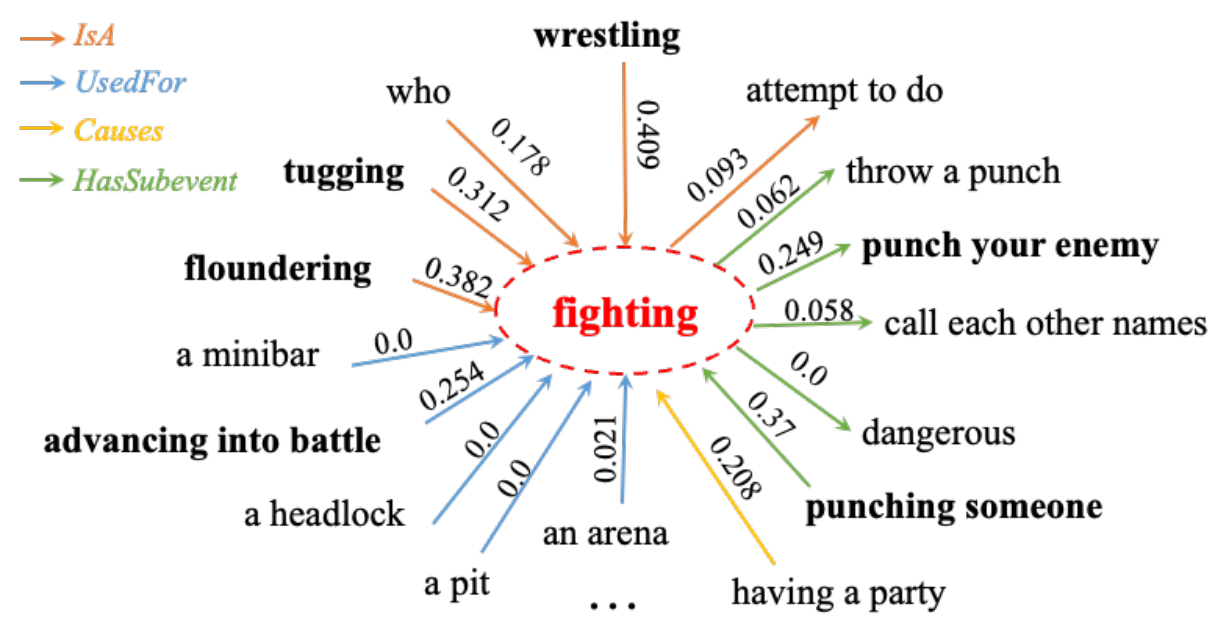}
\caption{An example of concept dictionary given the anomaly class \emph{fighting}. The arrows point from the head node to the tail node with relevance scores, and the colors indicate different relationships. The entire graph constitutes a concept dictionary, where the bold items are those retained after node filtering.}
\end{figure}

\subsubsection{Context Separation}
Since snippet-level features don't encapsulate complete contextual details, aligning them directly with prompt features could lead to inaccuracies in anomaly localization. Therefore, it's imperative to differentiate between class-specific foreground and class-agnostic background in successive snippets. We employ scaled anomaly scores as activations to generate video-level foreground and background features. These are represented as follows:

\begin{equation}
    {{\mathbf{A}}^{fg}}=\frac{\text{exp} \left( \mu {\mathbf{S}} \right)-1}{\mathop{\sum }_{t}\left(\text{exp} \left( \mu {\mathbf{S}_{t}} \right)-1 \right)},\text{ }{{\mathbf{V}}^{fg}}={{\mathbf{A}}^{fg}}{\mathbf{X}^{e}},
\end{equation}
\begin{equation}
    {{\mathbf{A}}^{bg}}=\frac{\text{exp} \left( \mu {{\bar{\mathbf{S}}}} \right)-1}{\mathop{\sum }_{t}\left( \text{exp} \left( \mu {{{\bar{\mathbf{S}}}}_{t}} \right)-1 \right)},\text{ }{{\mathbf{V}}^{bg}}={{\mathbf{A}}^{bg}}{{\mathbf{X}}^{e}},
\end{equation}
\noindent where $\mathbf{S}_t$ represents the anomaly score at the snippet-level, and $\mu$ is a predefined scaling factor that cooperates with the $\text{exp}(\cdot)$ operation to enhance high-confidence activations. In contrast, $\bar{\mathbf{S}}=1-\mathbf{S}_t$ represents the normal confidence of the current snippet, and $\mathbf{X}^e$ denotes the middle-layer output of the MLP, which shares the same dimension as $\mathbf{T}^c$.

\subsubsection{Cross-modal Alignment}
Finally, we present our approach to enhancing the fine-grained semantics of visual features by aligning them with the prompt features, as illustrated in Figure 2. For abnormal videos, foreground features $\mathbf{V}^{fg}_a$ are matched with the corresponding abnormal prompt $\mathbf{T}_a^c$, increasing the chances of accurate abnormal behavior identification. In contrast, background features $\mathbf{V}^{bg}_a$ are paired with the normal prompt $\mathbf{T}^c_n$, maintaining normal classification within an abnormal setting. For semantically inconsistent video-prompt pairs, repulsion is achieved by minimizing the cosine distance of all negative pairs, further forming discriminative decision boundaries. The process is formulated as:
\begin{equation}
    \psi\left( \mathbf{V},\mathbf{T} \right)=\frac{\mathbf{V}\cdot {{\mathbf{T}}}^\top}{\left\| \mathbf{V} \right\|\left\| \mathbf{T} \right\|},
\end{equation}
\begin{equation}
    p_{i}^{v2t}\left(\mathbf{V} \right)=\frac{\text{exp} \left(\psi\left(\mathbf{V},\mathbf{T} \right)/\tau  \right)}{\mathop{\sum }_{k=1}^{C+1}\text{exp} \left(\psi\left(\mathbf{V},{{\mathbf{T}}_{k}} \right)/\tau  \right)},
\end{equation}
\noindent where $\psi(\cdot)$ measures the cosine similarity between the visual representation $\mathbf{V}\in \left\{ \mathbf{V}_{a}^{fg}\bigcup \mathbf{V}_{a}^{bg}\bigcup \mathbf{V}_{n}^{fg} \right\}$ and the textual representation $\mathbf{T}\in \left\{ \mathbf{T}_{a}^{c}\bigcup \mathbf{T}_{n}^{c} \right\}$. Probability $p^{v2t}(\cdot)$ estimates how likely a visual feature matches a specific prompt across $C$ anomaly classes and 1 normal class, where $\tau$ is a temperature coefficient.
Finally, the cross-modal alignment loss is computed using the Kullback-Leibler divergence, which forces the network to learn to distinguish between the visual content of the video that represents abnormal behavior (foreground) and the content that is not relevant to the abnormal behavior (background). The loss function is formulated as follows:
\begin{equation}
    \mathscr{L}_{kd}=\mathbb{E}_{p \sim p(v)}[\text{log}\,p^{v2t}(v)-\text{log}\,q^{v2t}(v)],
\end{equation}
\noindent where $p^{v2t}(v)$ and $q^{v2t}(v)$ denote the similarity score and semantic consistency label of the video-prompt pair, respectively. If it is a positive pair, $q=1$; otherwise, $q=0$.

% \begin{figure}[!t]
% \centering
% \includegraphics[width=\linewidth]{method_fig/score_smooth.pdf}
% \caption{Illustration of Score Smoothing.}
% \end{figure}

\subsection{Training and testing procedures}
In the training phase, the overall objective function of our model is  denoted as:
\begin{equation}
    \mathscr{L}=\mathscr{L}_{ce} + \lambda\mathscr{L}_{kd},
\end{equation}
\noindent where the coefficient $\lambda$ is utilized to adjust the alignment loss. By optimizing this objective function, our model acquires the ability to generate discriminative representations of positive and negative snippets, while also effectively capturing the fine-grained semantics of anomalies. As a result, the generalizability of our model in complex situations is improved.

Considering the temporal consistency, we utilize a score smoothing (SS) strategy in the testing phase to mitigate the impact of transient noise that does not conform to the expected duration of an anomaly. This strategy employs distinct pooling operations to achieve optimal results. Given an anomaly score sequence $\{s_i\}_{i=1}^T$ and a pooling window of size $\kappa$, the smoothing process can be expressed as follows:
\begin{equation}
    \Tilde{s}_i=\frac{1}{\kappa}\underset{i=\kappa_s}{\overset{\kappa_e}{\mathop \sum }}\,s_i,
\end{equation}
\noindent where $\kappa_s$ and $\kappa_e$ denote the start and end positions of pooling, respectively. For a moving window, we set $\kappa_s=\left\lfloor {i}/{\kappa}\right\rfloor \kappa$ and $\kappa_e=\left( \left\lfloor {i}/{\kappa} \right\rfloor +1 \right)\kappa-1$. For a sliding window, we define $\kappa_s=i$ and $\kappa_e=i+\kappa-1$. In cases where the score sequence is shorter than the window size, the remainder is padded with zeroes. By smoothing the prediction scores, individual biases can be effectively suppressed while further reducing the occurrence of false alarms.

\section{Experiments}
In this section, we first introduce the datasets and implementation details of our framework. Subsequently, we make a comparison between our approach and state-of-the-art methods. Finally, we assess the contribution of each component via extensive ablation studies and qualitative analysis.

\subsection{Datasets and Evaluation Metric}
We perform experiments on three challenging anomaly benchmarks, i.e., UCF-Crime \cite{Sultani2018}, XD-Violence \cite{Wu2020}, and ShanghaiTech \cite{Luo2017} datasets. Details are given as follows:

\subsubsection{UCF-Crime}
It comprises 1900 surveillance videos with a total duration of 128 hours, covering 13 real-world anomalies, including abuse, robbery, explosion, and road accidents. For the weakly supervised setting, the dataset is divided into 1610 training and 290 test videos, with only video-level annotations available for training and frame-level annotations provided for testing.

\subsubsection{XD-Violence}
It is the latest and largest multimodal violence dataset containing 4754 untrimmed videos with a total duration of 217 hours. The dataset includes 3954 training and 800 test videos from various sources such as surveillance, movies, car cameras, and games. The dataset covers six types of violence, including abuse, car accidents, explosions, fighting, riots, and shooting. Most of the videos in this dataset contain artistic expressions such as camera movements and scene switching, which poses a challenge for video anomaly detection.

\subsubsection{ShanghaiTech}
It comprises 437 videos from 13 campus scenes. The original version was used for semi-supervised anomaly detection, where the training set consisted of only normal videos. Zhong \emph{et al}. \cite{Zhong2019} reorganized the dataset for weakly supervised setting, with 238 videos in the training set and 199 videos in the test set.

\subsubsection{Evaluation Metric}
For the UCF-Crime and ShanghaiTech datasets, we use the area under the frame-level receiver operating characteristic curve (AUC) as the evaluation metric. For the XD-Violence dataset, the area under the precision-recall curve, also known as the average precision (AP), is utilized as the standard evaluation metric. In addition, to guarantee the reliability of anomaly prediction, a false positive rate with a threshold value of 0.5, or false alarm rate (FAR), is employed for evaluation. In general, the lower the FAR, the more robust the model is to normal samples.

\subsection{Implementation Details}

\subsubsection{Data Pre-processing}
Following existing work \cite{Wu2021}, we utilize the I3D network \cite{Carreira2017} pre-trained on Kinetics-400 for feature extraction. For a fair comparison, we utilize a 10-crop augmentation strategy, consisting of the center, four corners, and their mirrored counterparts, for the UCF-Crime and ShanghaiTech datasets. In contrast, a 5-crop augmentation strategy, consisting of the center and four corners, is employed for the XD-Violence dataset as in \cite{Wu2020}.

\subsubsection{Hyperparameter settings}
The hidden dimension of the TCA module is set to 128, and the fusion weight $\alpha$ is initialized to 0.5. The two Conv1D layers of MLP have 512 and 300 nodes, respectively, both with a dropout rate of 0.1. For the UCF-Crime, XD-Violence, and ShanghaiTech datasets, we adopt local window sizes of 9, 9, and 5, respectively. The causal convolution kernel size $\Delta t$ is set to 9, 3 and 3, respectively, while the temperature coefficient $\tau$ is initialized to 0.09, 0.05, and 0.2, respectively. The scaling factor $\mu$ for context separation is set to 10, and the loss coefficient $\lambda$ is set to 1 for the UCF-Crime and XD-Violence datasets, while it is set to 9 for the ShanghaiTech dataset.

\subsubsection{Training and Test Details}
During the training phase, our model is trained using the Adam optimizer \cite{kingma2014adam} with a batch size of 128 and 50 epochs in total. The initial learning rate for the three datasets is set to $5\times10^{-4}$, with a cosine decay strategy. For the balance of computational efficiency and detection performance, the snippet sampling threshold in the training phase is set to 200 as in \cite{Wu2020} \cite{Wu2021} \cite{Zhen2021} \cite{Pu2022}. In the testing phase, we apply sliding pooling with window sizes of 7 and 3 to the UCF-Crime and ShanghaiTech datasets, respectively. Moving pooling with a window size of 9 is adopted for the XD-Violence dataset. More details can be found in our supplementary materials.

\subsection{Comparison With State-of-the-Art Methods}
We report the state-of-the-art results on the three benchmarks in Tables \uppercase\expandafter{\romannumeral1}-\uppercase\expandafter{\romannumeral3}. Notably, the weakly supervised-based methods exhibit superior performance compared to the semi-supervised learning methods across all datasets. The former, which solely relies on normal videos during the training phase, is susceptible to higher false alarms. This is primarily due to the weak generalization of these methods to unseen samples. On the other hand, the latter enables the model to learn discriminative patterns of both normal and abnormal samples, thus leading to better detection performance.

\begin{table}[!t]
\renewcommand\arraystretch{1.1}
\caption{Performance comparison of state-of-the-art methods on the UCF-Crime dataset}
\centering
\resizebox{\linewidth}{!}{
\begin{tabular}{c|cccc}
\toprule[1pt]
Supervision                                                                    & Method        & Feature  & AUC (\%)       & FAR (\%) \\ \hline
\multirow{3}{*}{\begin{tabular}[c]{@{}c@{}}Semi-\\ supervised\end{tabular}}    & Conv-AE \cite{Hasan2016}       & -        & 50.60           & 27.2     \\
                                                                               & Lu \emph{et al}. \cite{Lu2013}     & -  & 65.51          & 3.1      \\
                                                                               & GODS \cite{Wang2019}          & BoW+TCN  & 70.46          & 2.1      \\ \hline
\multirow{13}{*}{\begin{tabular}[c]{@{}c@{}}Weakly-\\ supervised\end{tabular}} & MIL-Rank \cite{Sultani2018}     & C3D RGB  & 75.41          & 1.9      \\
                                                                               & IBL \cite{Zhang2019}           & C3D RGB  & 78.66          & -        \\
                                                                               & Motion-Aware \cite{zhu2019motion}  & PWC Flow & 79.00             & -        \\
                                                                               & GCN  \cite{Zhong2019}   & TSN RGB  & 82.12          & \textbf{0.1}      \\
                                                                               & MIST \cite{Feng2021}         & I3D RGB  & 82.30           & {\ul0.13}     \\
                                                                               & HL-Net \cite{Wu2020}       & I3D RGB  & 82.44          & -        \\
                                                                               & MS-BSAD \cite{Zhen2021}      & I3D RGB  & 83.53          & -        \\
                                                                               & RTFM \cite{Tian2021}          & I3D RGB  & 84.30           & -        \\
                                                                                & CRFD \cite{Wu2021}          & I3D RGB  & 84.89           & 0.72        \\
                                                                               & DDL \cite{Pu2022}           & I3D RGB  & 85.12          & -     \\
                                                                               & MSL \cite{li2022self}          & I3D RGB  & 85.30           & -        \\
                                                                               & MLAD \cite{Zhang2022}         & I3D RGB  & 85.47          & 7.47     \\
                                                                               & NL-MIL \cite{Park2023}        & I3D RGB  & 85.63    & -        \\ 
                                                                               & S3R \cite{wu2022self}       & I3D RGB  & 85.99    & - 
                                                                               \\ 
                                                                               & Cho \emph{et al}. \cite{cho2023look}       & I3D RGB  & 86.10    & -   \\ 
                                                                               & CUPL \cite{zhang2023exploiting}       & I3D RGB  & 86.22    & -        \\ 
                                                                               & UML \cite{lv2023unbiased}       & X-CLIP RGB  & 86.75    & -        \\
                                                                               & UR-DMU \cite{zhou2023dual}       & I3D RGB  & \textbf{86.97}    & 1.05   \\
                                                                               \cline{2-5} 
                                                                               & \textbf{Ours} & I3D RGB  & {\ul 86.76} & 0.43     \\ \bottomrule[1pt]
\end{tabular}}
\end{table}

\begin{table}[!t]
\renewcommand\arraystretch{1.1}
\caption{Performance comparison of state-of-the-art methods on the XD-Violence dataset}
\centering
\begin{tabular}{c|cccc}
\toprule[1pt]
Supervision                                                                    & Method        & Feature       & AP (\%)        & FAR (\%)      \\ \hline
\multirow{3}{*}{\begin{tabular}[c]{@{}c@{}}Semi-\\ supervised\end{tabular}}    & SVM baseline          & -             & 50.78          & -             \\
                                                                               & OCSVM \cite{scholkopf1999support}        & -             & 27.25          & -             \\
                                                                               & Conv-AE \cite{Hasan2016}       & -             & 30.77          & -             \\ \hline
\multirow{12}{*}{\begin{tabular}[c]{@{}c@{}}Weakly-\\ supervised\end{tabular}} & MIL-Rank \cite{Sultani2018}      & C3D RGB       & 73.20           & -             \\
                                                                               & HL-Net \cite{Wu2020}       & I3D RGB       & 75.44          & -             \\
                                                                               & CA-VAD \cite{Chang2022}        & I3D RGB       & 76.90           & -             \\
                                                                               & RTFM \cite{Tian2021}         & I3D RGB       & 77.81
                                                                               & -             \\
                                                                               & CRFD \cite{Wu2021}         & I3D RGB  & 75.90           & -        \\
                                                                               & DDL \cite{Pu2022}           & I3D RGB       & 80.72    & -          \\
                                                                               & MSL \cite{li2022self}           & I3D RGB       & 78.28          & -             \\
                                                                               & NL-MIL \cite{Park2023}       & I3D RGB       & 78.51          & -             
                                                                               \\
                                                                               & S3R \cite{wu2022self}       & I3D RGB       & 80.26          & -    
                                                                               \\ 
                                                                               & UR-DMU \cite{zhou2023dual}       & I3D RGB  & {\ul 81.66}    & {\ul0.65}   
                                                                               \\ 
                                                                               & Cho \emph{et al}.\cite{cho2023look}       & I3D RGB  & 81.30    & - \\
                                                                               & ACF  \cite{wei2022look}       & I3D+VGGish  & 80.13    & - \\
                                                                               & MSAF  \cite{wei2022msaf}       & I3D+VGGish  & 80.51    & -     \\
                                                                               & CUPL \cite{zhang2023exploiting}       & I3D+VGGish  & 81.43    & -        \\ 
                                                                               & CMA-LA \cite{Pu2022cma}        & I3D+VGGish & 83.54          & - \\
                                                                               & MACIL-SD \cite{Yu2022}     & I3D+VGGish & 83.40           & -             \\ \cline{2-5} 
                                                                               & \textbf{Ours} & I3D RGB       & \textbf{85.59} & \textbf{0.57}          \\ \bottomrule[1pt]
\end{tabular}
\end{table}

\begin{table}[!t]
\renewcommand\arraystretch{1.1}
\caption{Performance comparison of state-of-the-art methods on the ShanghaiTech dataset}
\centering
\resizebox{\linewidth}{!}{
\begin{tabular}{c|cccc}
\toprule[1pt]
Supervision                                                                    & Method        & Feature       & AUC (\%)       & FAR (\%)   \\ \hline
\multirow{3}{*}{\begin{tabular}[c]{@{}c@{}}Semi-\\ supervised\end{tabular}}    & Mem-AE \cite{Gong2019}        & -             & 71.20          & -          \\
                                                                               & $\text{HF}^2$-VAD \cite{Liu2021}          & -             & 76.20          & -          \\
                                                                               & AMP-Net \cite{liu2023amp}          & -             & 78.80          & -          \\ \hline
\multirow{13}{*}{\begin{tabular}[c]{@{}c@{}}Weakly-\\ supervised\end{tabular}} & MIL-Rank \cite{Sultani2018}     & C3D RGB       & 86.30          & 0.15       \\
                                                                               & IBL \cite{Zhang2019}           & C3D RGB       & 82.50          & 0.10       \\
                                                                               & GCN  \cite{Zhong2019}   & TSN RGB       & 84.44          & -          \\
                                                                               & CLAWS \cite{zaheer2020claws}         & C3D RGB       & 89.67          & -          \\
                                                                               & AR-Net \cite{Wan2020}       & RGB+Flow & 91.24          & 0.10       \\
                                                                               & MIST \cite{Feng2021}          & I3D RGB       & 94.83          & {\ul 0.05} \\
                                                                               & CRFD \cite{Wu2021}         & I3D RGB       & 97.48          & -          \\
                                                                               & RTFM \cite{Tian2021}         & I3D RGB       & 97.21          & -          \\
                                                                               & MSL \cite{li2022self}           & VideoSwin & 97.32          & -          \\
                                                                               & NL-MIL \cite{Park2023}       & I3D RGB       & 97.43          & -          \\
                                                                               & S3R \cite{wu2022self}          & I3D RGB       & 97.48          & -          \\
                                                                               & UML \cite{lv2023unbiased}          & X-CLIP RGB    & 96.78          & - \\ 
                                                                               & Cho \emph{et al}.\cite{cho2023look}       & I3D RGB  & {\ul 97.60}    & -          \\ \cline{2-5} 
                                                                               & \textbf{Ours} & I3D RGB       & \textbf{98.14} & \textbf{0.00}       \\ \bottomrule[1pt]
\end{tabular}}
\end{table}

\begin{table}[!t]
\renewcommand\arraystretch{1.1}
\caption{The contribution of each component. TCA: temporal context aggregation. PEL: prompt-enhanced learning. SS: score smoothing.\label{tab:table1}}
\centering
\begin{tabular}{cccc|ccc}
\toprule[1pt]
Baseline & TCA & PEL & SS & \begin{tabular}[c]{@{}c@{}}UCF\\ AUC (\%)\end{tabular} & \begin{tabular}[c]{@{}c@{}}XD\\ AP (\%)\end{tabular} & \begin{tabular}[c]{@{}c@{}}SHTech\\ AUC (\%)\end{tabular} \\ \hline
\ding{51}        & \ding{55}    &  \ding{55}   &  \ding{55}  & 82.50                                                        & 73.51                                                         & 92.25                                                           \\
\ding{51}        & \ding{51}   &  \ding{55}   &  \ding{55}  & 85.72                                                        & 83.28                                                         & 97.82                                                           \\
\ding{51}        & \ding{51}   & \ding{51}   &  \ding{55}  & 86.36                                                        & 85.26                                                         & 98.00                                                           \\
\ding{51}        & \ding{51}   & \ding{51}   & \ding{51}  & \textbf{86.76}                                                        & \textbf{85.59}                                                         & \textbf{98.14}                                                           \\ \bottomrule[1pt]
\end{tabular}
\end{table}

Our model achieves competitive results across three datasets among all weakly supervised methods. Although our model's AUC value on the UCF-Crime dataset is lower than that of UR-DMU \cite{zhou2023dual} by 0.21\%, we outperform UR-DMU on the XD-Violence dataset with an absolute gain of 3.93\% AP and lower false alarm rates on both datasets. UR-DMU employs an additional encoder layer to learn local features with temporal masks, whereas our TCA module achieves local-global context modeling by reusing the similarity matrix. Additionally, UR-DMU introduces a dual memory unit to store normal and abnormal patterns and enlarges the decision boundary through a magnitude distance loss. In contrast, our proposed PEL module enhances fine-grained detection performance by considering the intra-class discriminability of exceptions while ensuring inter-class separability. Notably, our model even outperforms multimodal-based approaches, namely CMA-LA \cite{Pu2022cma} and MACIL-SD \cite{Yu2022}, with 2.19\% and 2.05\% gains on the XD-violence dataset, respectively. Although the PEL module requires extra textual information in the training phase, our model only requires appearance features as input during inference. Furthermore, our model achieves a new state-of-the-art performance on the ShanghaiTech dataset, with a 0.54\% improvement in AUC compared to Cho \emph{et al} \cite{cho2023look}. For the first time, the false alarm rate is also reduced to 0, indicating that the model has sufficient confidence and robustness for normal samples.

\subsection{Ablation Studies}
In this subsection, we perform extensive ablation studies to verify the contribution of each component of our model.

\subsubsection{Contribution of Each Component}
% Table \uppercase\expandafter{\romannumeral4} presents the contributions of each component, where `Baseline' refers to the combination of MLP and CC module. Significantly, the TCA module is responsible for the primary improvement, delivering AUC gains of 9.77\% and 5.57\% on the ShanghaiTech and XD-Violence datasets, respectively, highlighting the need for context modeling. Furthermore, the introduction of the PEL module enhances the detection performance of our model on all three datasets, with a substantial AP gain of 1.98\% on XD-Violence. We contend that this is due to the internal distribution characteristics of the dataset, which has a wider range of acquisition sources than surveillance videos such as UCF-Crime and ShanghaiTech, resulting in diverse and complex video content. Finally, the SS strategy further improves the detection performance of our model by 0.4\%, 0.33\%, and 0.14\% on the three benchmarks, respectively, demonstrating the effectiveness and generality of the module.

Table \uppercase\expandafter{\romannumeral4} details the impact of each model component. The `Baseline', comprising MLP and the Classifier, is notably enhanced by the TCA module, which yields substantial improvements of 9.77\% and 5.57\% on XD-Violence and ShanghaiTech datasets, respectively. This underscores the importance of context modeling. The PEL module further advances detection across all datasets, particularly notable with a 1.98\% AP increase on XD-Violence. This reflects the varied content sources of XD-Violence compared to UCF-Crime and ShanghaiTech. The SS strategy also demonstrates consistent, albeit smaller, improvements across the benchmarks, confirming its effectiveness and versatility.

\begin{table}[!t]
\renewcommand\arraystretch{1.1}
\caption{Contribution of Internal Components of the TCA module. DPE: Dynamic Position Encoding.}
\centering
\resizebox{\linewidth}{!}{
\begin{tabular}{ccc|ccc}
\toprule[1pt]
Local & Global & DPE & \begin{tabular}[c]{@{}c@{}}UCF\\ AUC (\%)\end{tabular} & \begin{tabular}[c]{@{}c@{}}XD\\ AP (\%)\end{tabular} & \begin{tabular}[c]{@{}c@{}}SHTech\\ AUC (\%)\end{tabular} \\ 
\midrule % Use \midrule instead of \hline for booktabs
\ding{51} & \ding{55}   &  \ding{55}   &  84.64 & 77.29 & 92.79 \\
\ding{55} & \ding{51}   &  \ding{55}   &  82.30 & 80.80 & 85.65 \\
\ding{51} & \ding{51}   & \ding{55}   &  85.18 & 83.05 & 97.62 \\
\ding{51} & \ding{51}   & \ding{51}   & \textbf{85.72} & \textbf{83.28} & \textbf{97.82} \\ 
\bottomrule[1pt]
\end{tabular}}
\end{table}

% \usepackage{multirow}
\begin{table}[!t]
\renewcommand\arraystretch{1.1}
\caption{The superiority of adaptive fusion in the TCA module.}
\centering
\begin{tabular}{ccc|ccc}
\toprule[1pt]
Weight & Global & Local & \begin{tabular}[c]{@{}c@{}}UCF\\ AUC (\%)\end{tabular} & \begin{tabular}[c]{@{}c@{}}XD\\ AP (\%)\end{tabular} & \begin{tabular}[c]{@{}c@{}}SHTech\\ AUC (\%)\end{tabular} \\ 
\midrule
\multirow{5}{*}{Fixed} & 0.1 & 0.9 & 84.82 & 77.57 & 97.74 \\
                       & 0.3 & 0.7 & 85.07 & 81.08 & 97.57 \\
                       & 0.5 & 0.5 & 85.17 & 82.45 & 97.71 \\
                       & 0.7 & 0.3 & 85.24 & 82.53 & 97.56 \\
                       & 0.9 & 0.1 & 84.85 & 80.57 & 97.60 \\                                           
\midrule
Learnable & $\alpha$ & $1-\alpha$ & \textbf{85.72} & \textbf{83.28} & \textbf{97.82} \\ 
\bottomrule[1pt]
\end{tabular}
\end{table}

\begin{table}[!t]
\renewcommand\arraystretch{1.1}
\caption{Comparison of different normalization methods for adaptive fusion}
\centering
\begin{tabular}{l|ccc}
\toprule[1pt]
Normalization   & \begin{tabular}[c]{@{}c@{}}UCF\\ AUC (\%)\end{tabular} & \begin{tabular}[c]{@{}c@{}}XD\\ AP (\%)\end{tabular} & \begin{tabular}[c]{@{}c@{}}SHTech\\ AUC (\%)\end{tabular} \\ \hline
None            & 83.44                                                  & \textbf{83.28}                                                & 94.06                                                     \\
Power Norm      & 82.88                                                  & 79.04                                                & 94.52                                                     \\
L2 Norm         & 85.09                                                  & 73.92                                                & 97.48                                                     \\
Power + L2 Norm & \textbf{85.72}                                                  & 71.13                                                & \textbf{97.82}                                                     \\ \bottomrule[1pt]
\end{tabular}
\end{table}

\subsubsection{Contribution of Components of the TCA module}
% Table \uppercase\expandafter{\romannumeral5} demonstrates that local context modeling can effectively improve the detection performance of our baseline. However, global context modeling resulted in a significant degradation of performance on the UCF-Crime and ShanghaiTech datasets, while outperforming local context modeling on XD-Violence. We assert that global context modeling introduces long-range noise to fixed surveillance videos, leading to reduced discrimination among snippet features. In contrast, film and TV videos require a global receptive field to capture complex semantics across snippets. The adaptive fusion of local and global contexts significantly enhanced the model's performance, indicating that each approach plays a complementary role to the other, which outperformed the baseline by 9.54\% and 5.37\% on the XD-Violence and ShanghaiTech datasets, respectively. Finally, the improvement from DPE indicates the importance of relative distance as a location prior.

Table \uppercase\expandafter{\romannumeral5} shows that local context modeling significantly improves baseline detection. In contrast, global context modeling decreases performance on UCF-Crime and ShanghaiTech but is more effective on XD-Violence. We argue that global context may introduce noise in fixed surveillance footage, reducing feature discrimination, but is beneficial for diverse film and TV content. Combining local and global contexts outperforms the baseline by 9.54\% and 5.37\% on XD-Violence and ShanghaiTech, respectively. This indicates the complementary roles of these approaches. The improvement from DPE highlights the importance of relative distance in feature location.

Furthermore, Table \uppercase\expandafter{\romannumeral6} demonstrates adaptive fusion's superiority over fixed weights, showing better results across all datasets. This supports the dynamic adjustment of local-global context fusion to suit diverse videos. Table \uppercase\expandafter{\romannumeral7} compares normalization methods. Power and L2 normalization combined yield optimal results on UCF-Crime and ShanghaiTech. However, on XD-Violence, either normalization method significantly reduces performance, likely affecting the dataset's diverse video source distribution. Consequently, we omit normalization for XD-Violence in adaptive fusion to preserve content diversity.

Finally, we visualize the correlation of context features to demonstrate the encoding performance of the TCA module. As shown in Figure 4, the heatmap demonstrates the cosine similarity between the snippets. From this, we observe: 1) UCF-Crime videos show less variability compared to XD-Violence due to their fixed surveillance nature versus XD-Violence's complex scene and camera movements; 2) Local features offer more discriminability than global features, which tend to introduce more long-range noise; and 3) The aggregation of local and global features suppresses redundancies and enhances snippet discriminability.

% Compared with fixed weights, the learned weights yield better results across all three datasets, which enables the dynamic adjustment of the fusion ratio of the local-global context to suit diverse datasets. Table \uppercase\expandafter{\romannumeral7} compares the performance of different normalization methods. The combination of power normalization and L2 normalization achieves the best results on UCF-Crime and ShanghaiTech. However, we observe that either power normalization or L2 normalization leads to a significant drop in performance for XD-Violence. We argue that normalization may jeopardize the diversity distribution of this dataset since it contains videos from different sources and types. While normalization is necessary for surveillance videos with fixed lenses, we remove the normalization operation in adaptive fusion for XD-Violence.

%%%%%%%%%%

\begin{figure}[!t]
\centering
\includegraphics[width=\linewidth]{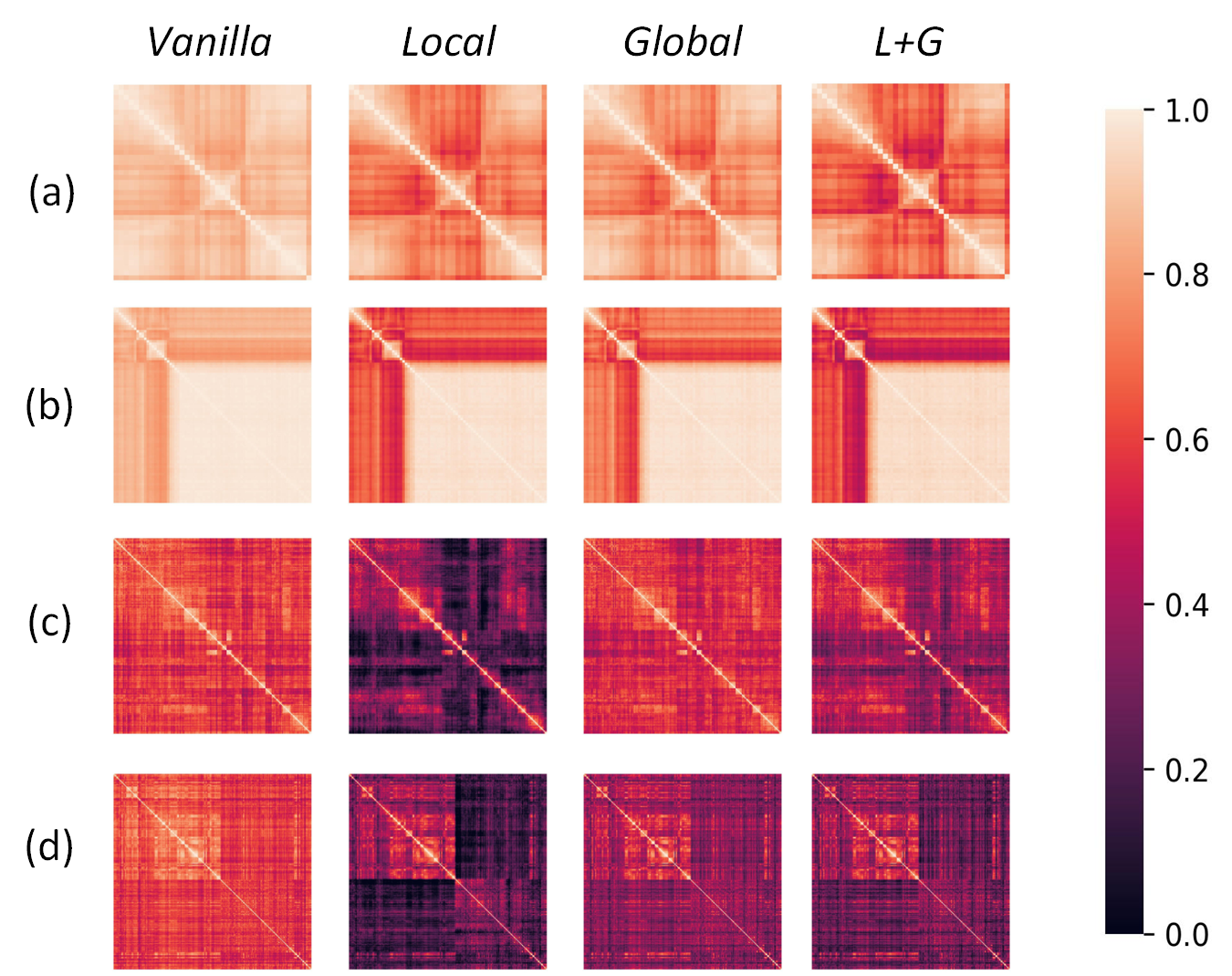}
\caption{Visualization of cosine similarity of context features. (a) and (b) are videos from the UCF-Crime dataset. (c) and (d) are videos from the XD-Violence dataset.}
\end{figure}

\begin{table}[!t]
\renewcommand\arraystretch{1.2}
\caption{Comparison of the spatio-temporal complexity of different methods}
\centering
\begin{threeparttable}
\begin{tabular}{ccc|cc}
\toprule[1pt]
Method       & \# Params & \# FLOPS  & \begin{tabular}[c]{@{}c@{}}UCF-Crime\\ AUC(\%)\end{tabular} & \begin{tabular}[c]{@{}c@{}}XD-Violence\\ AP(\%)\end{tabular} \\ \hline
$\text{HL-Net}^\dag$ \cite{Wu2020}     & 0.66M  & 133M   & 83.12                                                 & 79.58                                               \\
$\text{MTCM}^\dag$ \cite{Zhen2021}       & 13.27M & 2.653G & 83.15                                                 & 78.92                                               \\
$\text{RTFM}^\dag$ \cite{Tian2021}       & 24.72M & 791M   & 84.30                                                  & 77.81                                               \\
LA-Net \cite{Pu2022}     & 2.69M  & 538M   & 83.67                                                 & 79.18                                               \\
$\text{LGTE}^\dag$ \cite{Qing2021}       & 1.21M  & 241M   & 84.34                                                 & 79.07                                               \\ \hline
\textbf{TCA} & 1.21M  & 241M   & \textbf{85.72}                                        & \textbf{83.28}                                      \\ \bottomrule[1pt]
\end{tabular}
\begin{tablenotes}
        \footnotesize
        \item[\dag] refers to re-implementation in our baseline. 
      \end{tablenotes}

\end{threeparttable}
\end{table}

\subsubsection{Analysis of Time–Space Complexity}

This subsection evaluates the Temporal Context Aggregation (TCA) module's performance and efficiency against recent temporal modeling techniques, shown in Table \uppercase\expandafter{\romannumeral8}. For a fair comparison, we used open-source codes to reproduce existing modules and integrated them into our baseline. HL-Net \cite{Wu2020} uses dual graph convolutional networks for global and local relationship modeling. MTCM \cite{Zhen2021} employs multiple 1D convolution layers for multi-scale context, and RTFM \cite{Tian2021} uses self-attention with pyramid dilated convolution for understanding temporal relations. These methods significantly increase parameters and computational demands. However, our TCA module, with similar space-time complexity, outperforms LGTE \cite{Qing2021} by 1.38\% and 4.21\% on UCF-Crime and XD-Violence datasets, respectively. The TCA module's success is due to its effective use of the similarity matrix, maintaining contextual continuity. This contrasts with LGTE's channel grouping, which might lose some feature channel details. TCA's adaptive aggregation and dynamic positional coding also improve snippet feature uniqueness. Notably, the feature extraction model I3D can achieve a processing speed of 263 FPS on a single Tesla A40 GPU when the input frame resolution is set to \(224\times 224\). Subsequently, the anomaly detection model outputs anomaly scores at 350 FPS (2.9 ms per frame), striking a good balance between detection performance and efficiency.

\begin{table}[!t]
\renewcommand\arraystretch{1.1}
\caption{Contribution of node filtering in prompt construction}
\centering
\begin{tabular}{l|cc}
\toprule[1pt]
Method            & \begin{tabular}[c]{@{}c@{}}UCF-Crime\\ AUC (\%)\end{tabular} & \begin{tabular}[c]{@{}c@{}}XD-Violence\\ AP (\%)\end{tabular} \\ \hline
None     & 85.34                                                  & 84.35                                                \\
step\,1          & 85.87                                                  & 84.61                                                \\
step\,1 + step\,2 (fixed threshold) & 86.31                                                  & 85.10                                                 \\
step\,1 + step\,2 (dynamic threshold) & \textbf{86.36}                                         & \textbf{85.26}                                       \\ \bottomrule[1pt]
\end{tabular}
\end{table}

\begin{table}[!t]
\renewcommand\arraystretch{1.1}
\caption{performance comparison of different prompt template}
\centering
\begin{tabular}{l|cc}
\toprule[1pt]
Prompt Template                 & \begin{tabular}[c]{@{}c@{}}UCF-Crime\\ AUC (\%)\end{tabular} & \begin{tabular}[c]{@{}c@{}}XD-Violence\\ AP (\%)\end{tabular} \\ \hline
\{label\}                       & 85.95                                                        & 84.69                                                         \\
\{`a video of' + label\}          & 85.55                                                        & 84.51                                                         \\
\{`a long video of' + label\}     & 85.63                                                        & 84.59                                                         \\
\{label + WordNet Definition\} \cite{yao2022detclip}  & 85.92                                                        & 84.14                                                         \\ 
\{Learnable Prompt + label\} \cite{ju2022prompting} & 85.48                                               & 84.48    \\ \hline
\{label + ConceptNet Relation\} & \textbf{86.36}                                               & \textbf{85.26}                                            \\ \bottomrule[1pt]
\end{tabular}
\end{table}

\subsubsection{Contribution of Components of the PEL}
Table \uppercase\expandafter{\romannumeral9} first evaluates the impact of node filtering in prompt construction. Step 1 involves removing entries with relevance scores at or below zero, while Step 2 applies a specific threshold to the remaining entries. Step 1 enhances detection by 0.53\% and 0.26\% on UCF-Crime and XD-Violence datasets, respectively. Step 2 with a dynamic threshold further improves performance on both datasets, indicating effective reduction of redundant entries. The dynamic threshold maintains relevant semantic relationships, while a fixed threshold might include low-relevance, potentially anomalous entries.

\paragraph{Prompt Templates Comparison}
Different prompt templates are compared in Table \uppercase\expandafter{\romannumeral10}. Category labels as prompts increase detection by 0.23\% and 1.41\% on UCF-Crime and XD-Violence, respectively, surpassing hand-crafted templates and WordNet definitions. ConceptNet-based prompts further improve performance by 0.41\% and 0.57\% on these datasets. Using class-specific concepts based on semantic relations aids in identifying co-occurring anomalous items. Threshold filtering ensures the selection of highly relevant semantic concepts, enhancing fine-grained visual information capture.

\begin{table}[!t]
\renewcommand\arraystretch{1.1}
\caption{contribution of context separation in the PEL module}
\centering
\begin{tabular}{l|cc}
\toprule[1pt]
Method             & \begin{tabular}[c]{@{}c@{}}UCF-Crime\\ AUC (\%)\end{tabular} & \begin{tabular}[c]{@{}c@{}}XD-Violence\\ AP (\%)\end{tabular} \\ \hline
w/o context separation & 85.71                                                  & 79.79                                                \\
w/ context separation & \textbf{86.36}                                         & \textbf{85.26}                                       \\ \bottomrule[1pt]
\end{tabular}
\end{table}

\begin{table}[!t]
\renewcommand\arraystretch{1.1}
\caption{contribution of score smoothing strategy}
\centering
\begin{tabular}{l|ccc}
\toprule[1pt]
Method                 & \begin{tabular}[c]{@{}c@{}}UCF\\ AUC (\%)\end{tabular} & \begin{tabular}[c]{@{}c@{}}XD\\ AP (\%)\end{tabular} & \begin{tabular}[c]{@{}c@{}}SHTech\\ AUC (\%)\end{tabular} \\ \hline
None                 & 86.36                                                  & 85.26                                                & 98                                                        \\
moving window  & 86.65                                                  & \textbf{85.59}                                       & 98.03                                                     \\
sliding window & \textbf{86.76}                                         & 85.56                                                & \textbf{98.14}                                            \\ \bottomrule[1pt]
\end{tabular}
\end{table}

\paragraph{Context Separation}
As shown in Table \uppercase\expandafter{\romannumeral11}, `w/o context separation’ denotes the mean pooling operation of snippet features and alignment with the corresponding prompt features, leading to notable performance drops. Mean pooling dilutes the anomalous region's impact and introduces background noise, skewing cross-modal alignment. Therefore, context separation effectively minimizes background noise, ensuring better semantic consistency between visual foreground and prompt representations.

\subsubsection{PEL's Impact on Fine-Grained Anomaly Detection} As shown in Figure 5(a), our PEL method demonstrates superior anomaly detection capabilities on the UCF-Crime dataset, outperforming RTFM \cite{Tian2021} and UMIL \cite{lv2023unbiased} in most categories with significant leads in \emph{Abuse} (76.9\%) and \emph{Assault} (96.2\%). Its robust average AUC of 72.2\% underscores its effectiveness and potential applicability in security and surveillance. While PEL excels in complex anomaly scenarios, it lags slightly in \emph{Explosion}, suggesting a need for refinement in detecting abrupt anomalies.
In Figure 5(b), the performance of the PEL method in detecting various sub-classes of violence is contrasted with a baseline (w/o PEL). PEL consistently improves detection across all categories, with significant AP gains in \emph{Fighting} (83.8\% vs. 81.0\%) and \emph{Abuse} (70.5\% vs. 61.5\%). These improvements suggest that PEL effectively captures the nuances of complex, dynamic events. The overall average performance increase to 70.3\% from 68.1\% also underscores PEL's robustness across diverse scenarios. However, in \emph{Car Accident}, the impact of PEL is minimal, indicating potential areas for refinement.

\subsubsection{Impact of Score Smoothing}
Table \uppercase\expandafter{\romannumeral12} illustrates the effects of different pooling window sizes in the SS (Sliding and Sliding) strategy. Implementing both moving and sliding pooling techniques has shown to not only improve the model's performance but also decrease the false alarm rate. Specifically, the sliding pooling technique yields better results for ShanghaiTech and UCF-Crime datasets. Conversely, the moving pooling technique is more effective for the XD-Violence dataset.

\begin{figure}[!t]
\centering
\subfloat[Class-wise AUC results of three methods on UCF-Crime.]{\includegraphics[width=0.9\linewidth]{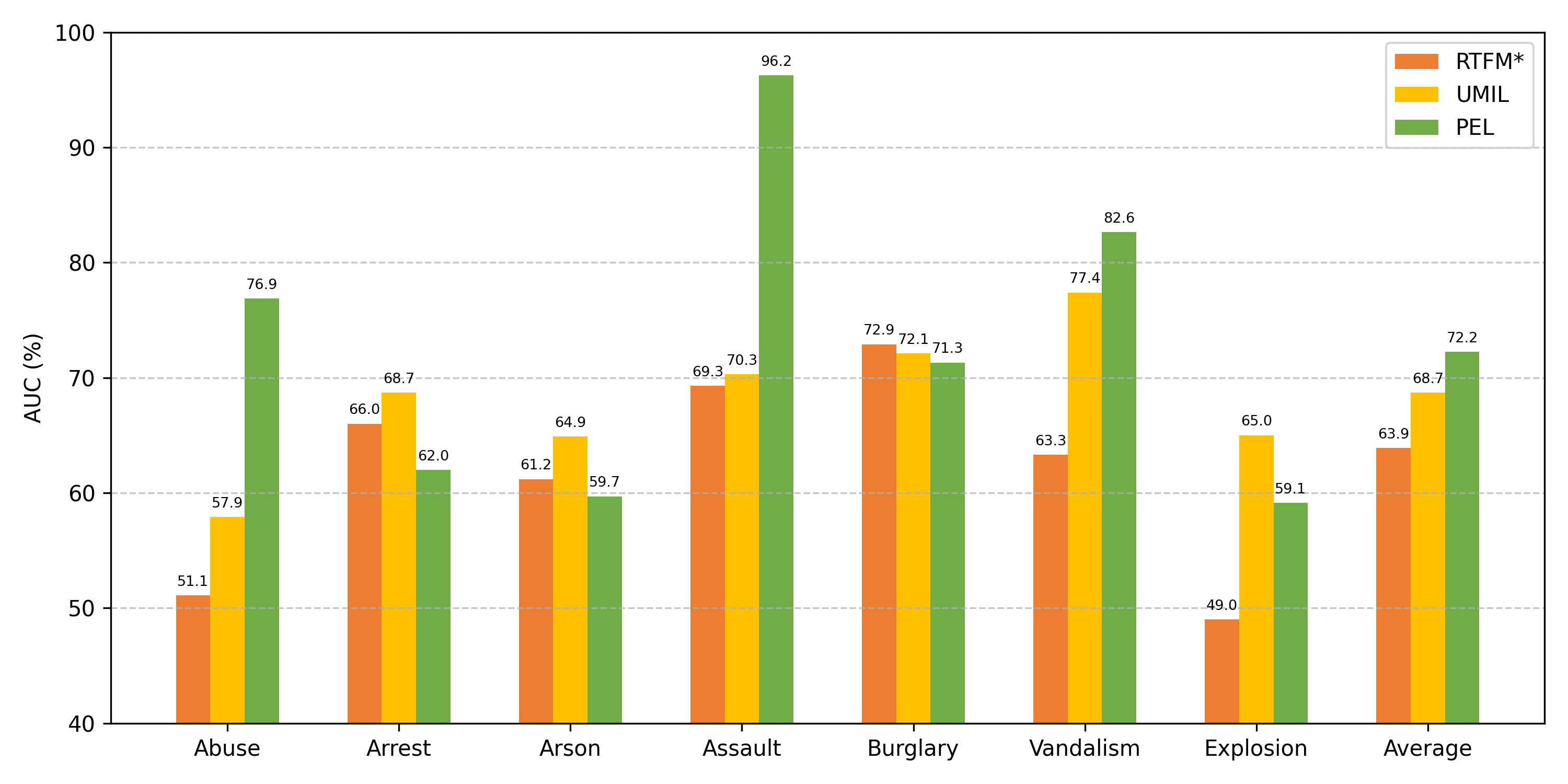}%
\label{fig_first_case}}
\hfil
\subfloat[Class-wise AP results of before/after PEL on XD-Violence.]{\includegraphics[width=0.9\linewidth]{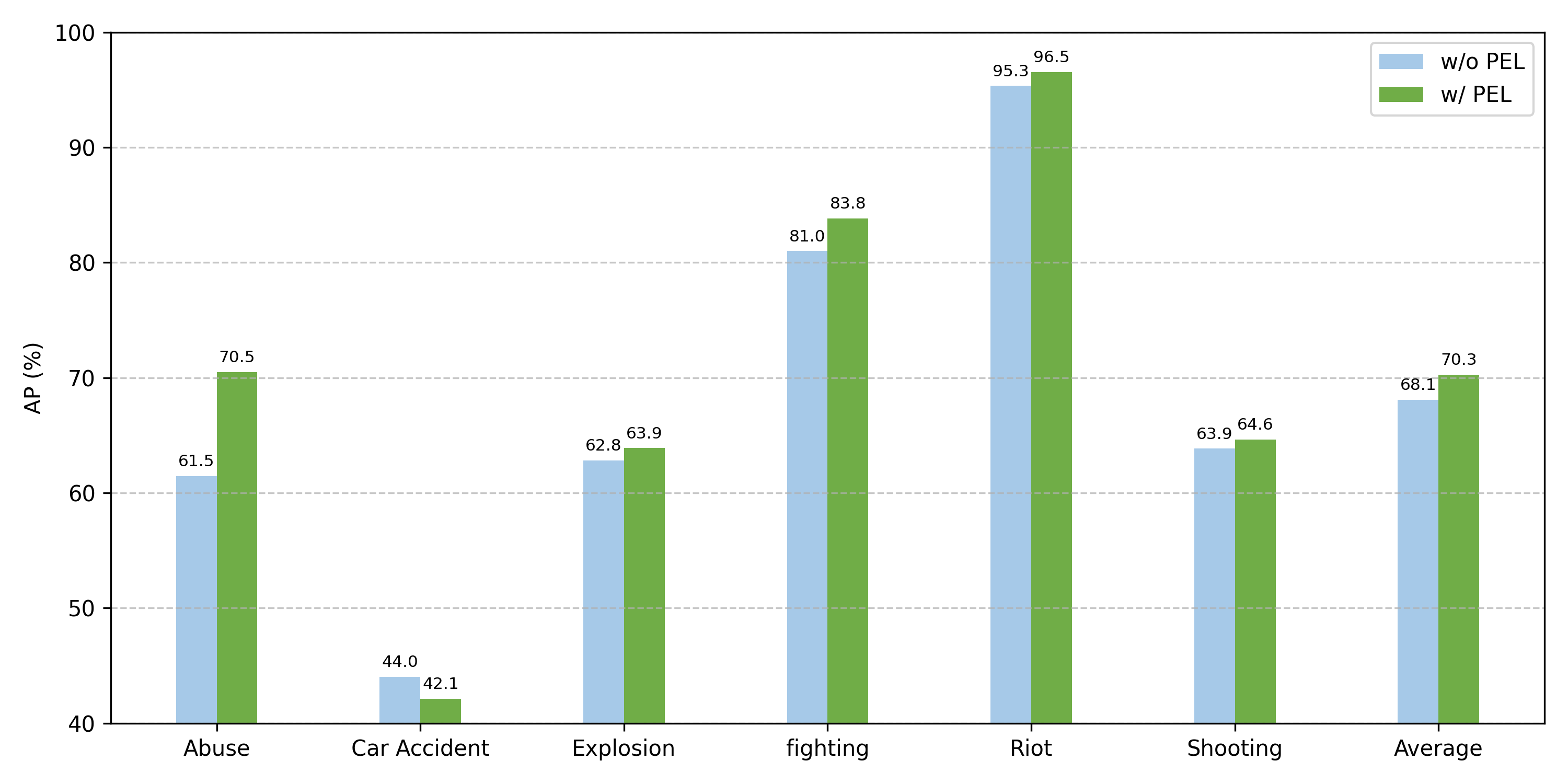}%
\label{fig_second_case}}
\caption{Contribution of the PEL module to fine-grained anomaly detection.}
\label{fig_sim}
\end{figure}

% \begin{figure*}[!h]
% \centering
% \subfloat[Performance versus $w$ when $\Delta t=9$.]{\includegraphics[width=0.49\linewidth]{window.png}%
% \label{fig_first_case}}
% \hfil
% \subfloat[Performance versus $\Delta t$ when $w=9$.]{\includegraphics[width=0.49\linewidth]{kernel.png}%
% \label{fig_second_case}}
% \caption{Changes of AUC and FAR w.r.t $w$ and $\Delta t$ on the UCF-Crime dataset.}
% \label{fig_sim}
% \end{figure*}

% \begin{figure*}[!t]
% \centering
% \subfloat[AUC results w.r.t. pooling type and window size on UCF-Crime.]{\includegraphics[width=0.49\linewidth]{UCF-SS.png}%
% \label{fig_first_case}}
% \hfil
% \subfloat[AP results w.r.t. pooling type and window size on XD-Violence.]{\includegraphics[width=0.49\linewidth]{XD-SS.png}%
% \label{fig_second_case}}
% \caption{Performance comparison of different pooling types and window sizes in SS strategy.}
% \label{fig_sim}
% \end{figure*}

\begin{figure*}[!t]
\centering
\subfloat[]{\includegraphics[width=0.25\linewidth]{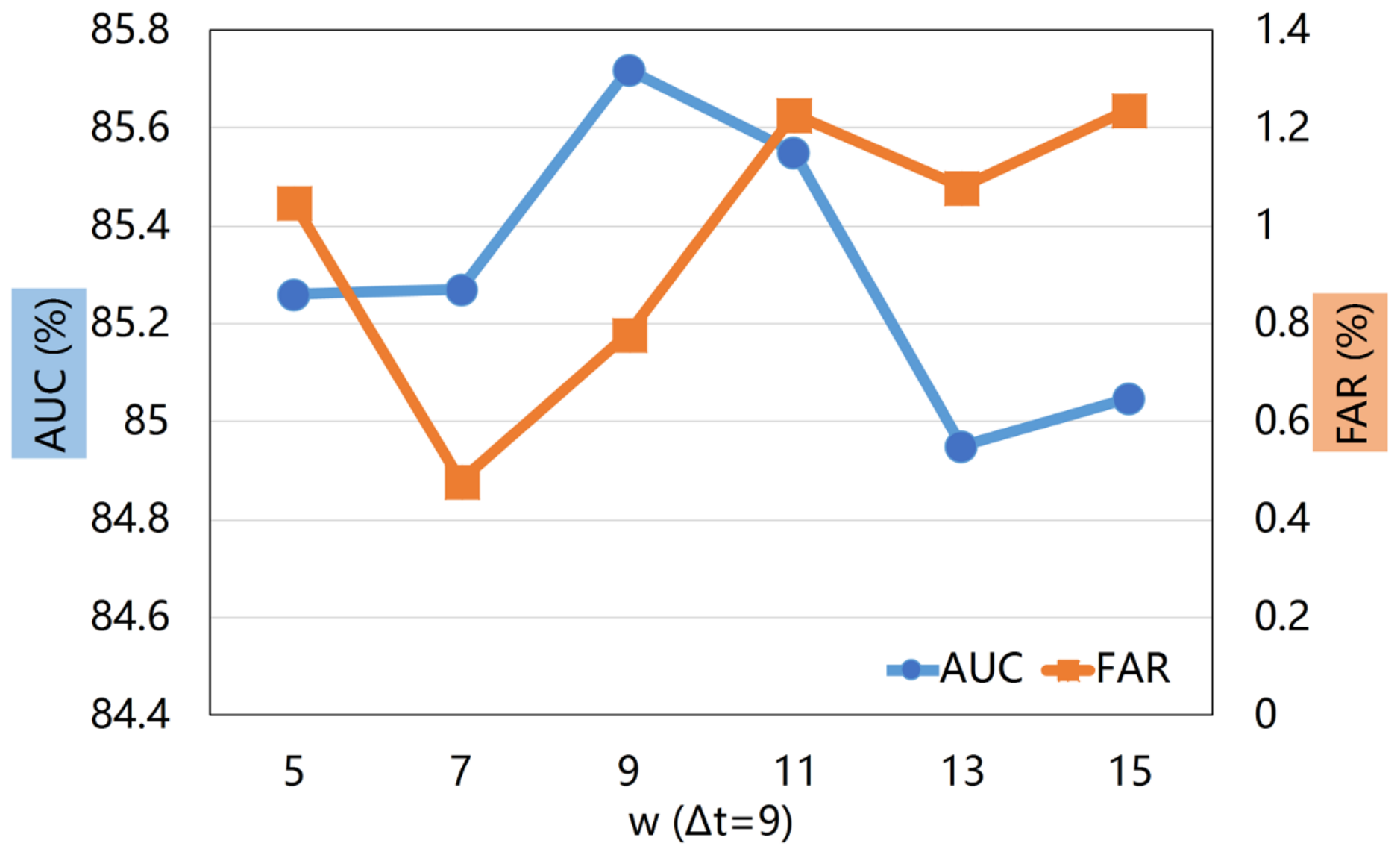}}
\hfil
\subfloat[]{\includegraphics[width=0.25\linewidth]{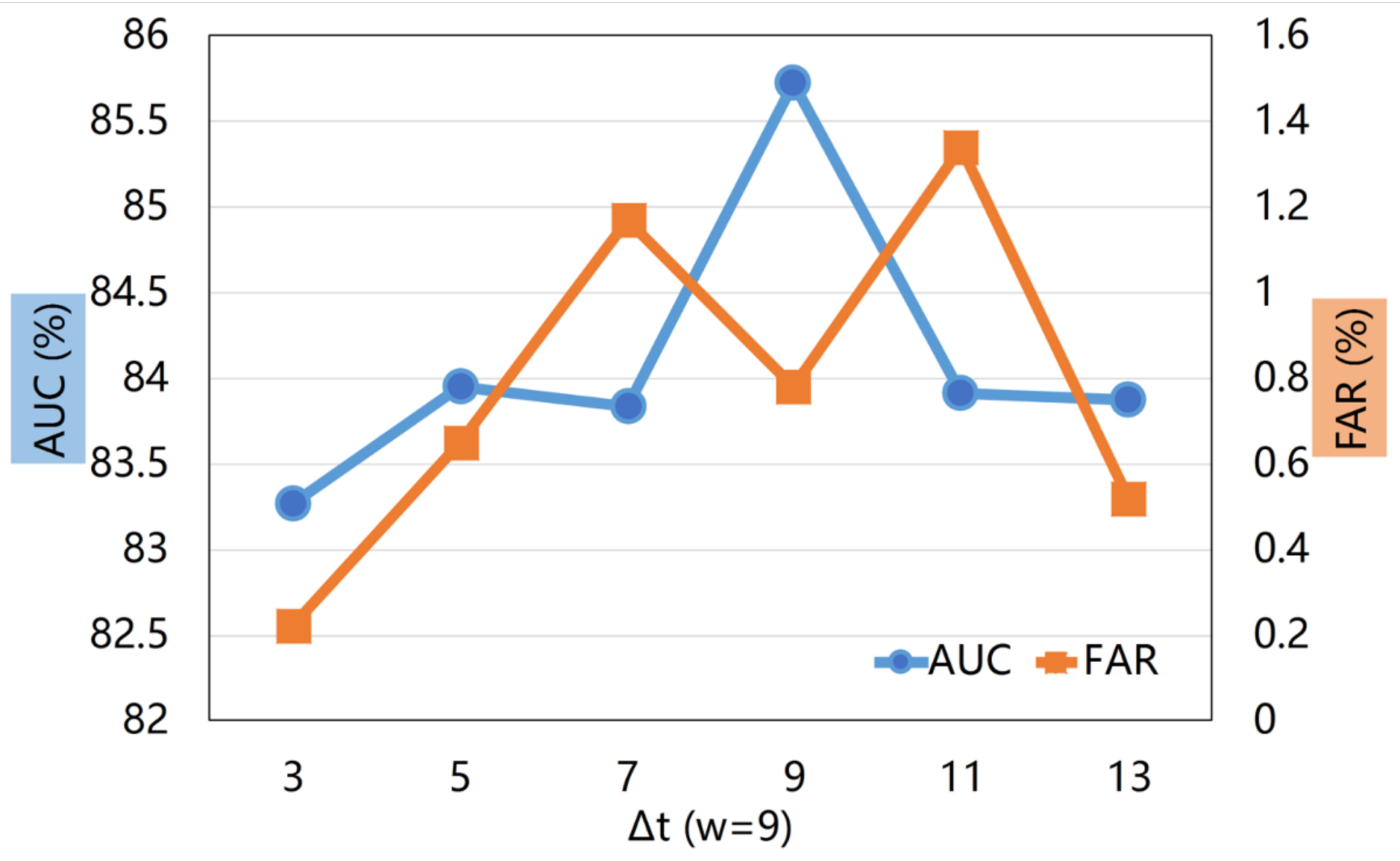}}
\hfil
\subfloat[]{\includegraphics[width=0.24\linewidth]{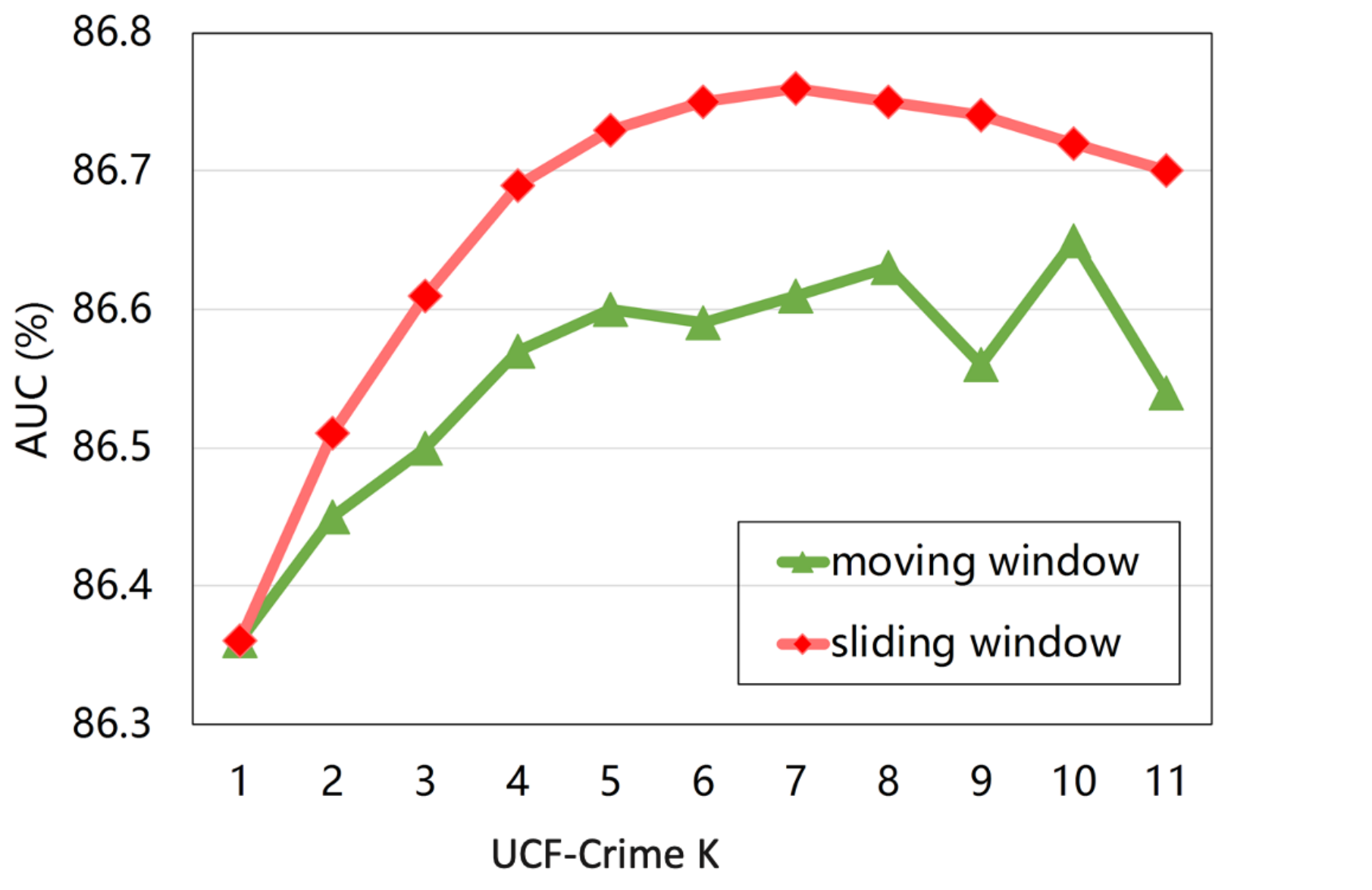}}
\hfil
\subfloat[]{\includegraphics[width=0.24\linewidth]{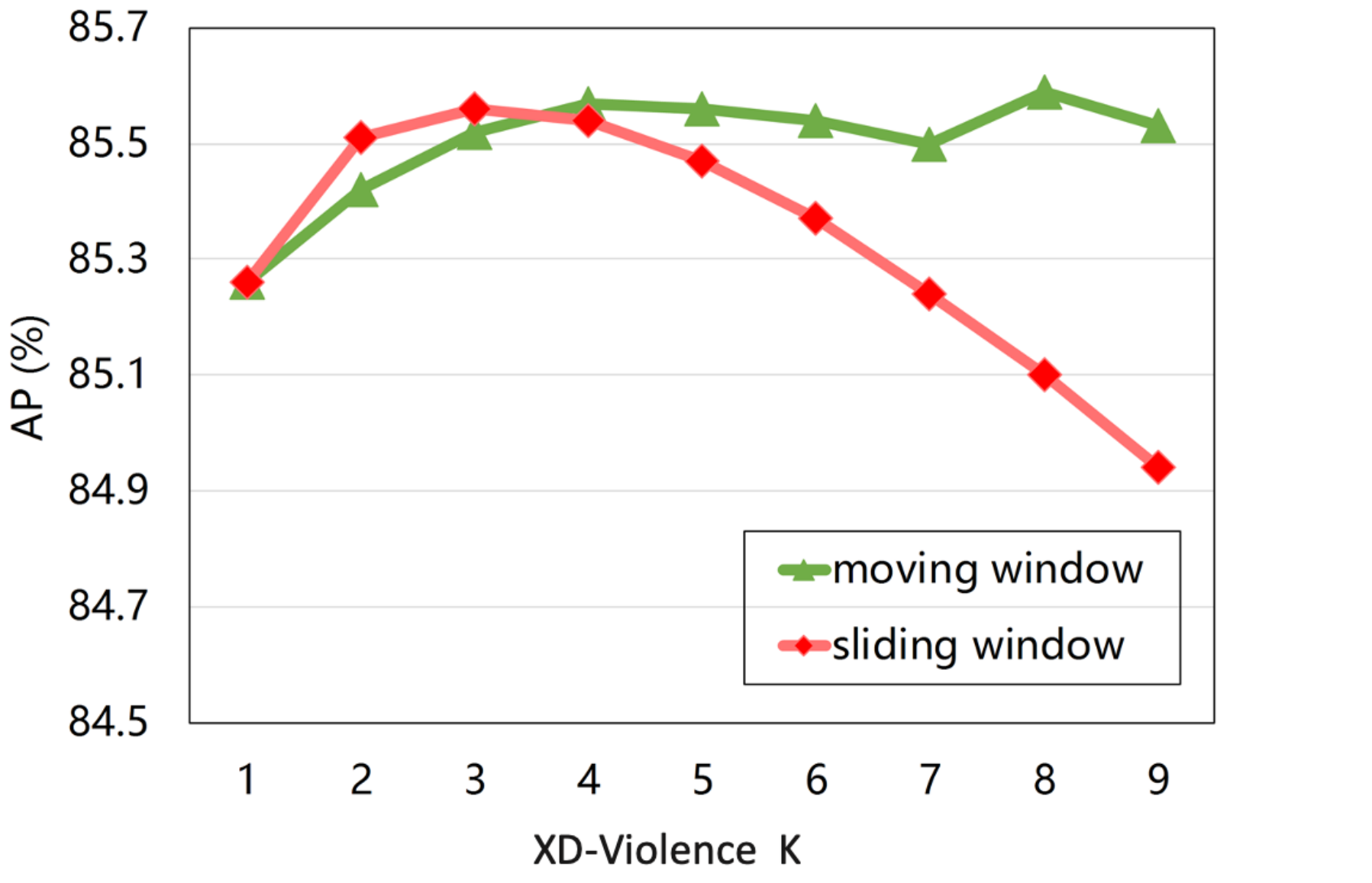}}
\caption{Model performance under different hyperparameter settings. (a) and (b) show the AUC and FAR results of our model for different local windows $w$ and causal convolution size $\Delta t$, respectively. (c) and (d) illustrate the effects of two different score smoothing methods on the model's performance as the window size varies.}
\end{figure*}

\begin{figure*}[!t]
\centering
\subfloat[]{\includegraphics[width=0.5\linewidth]{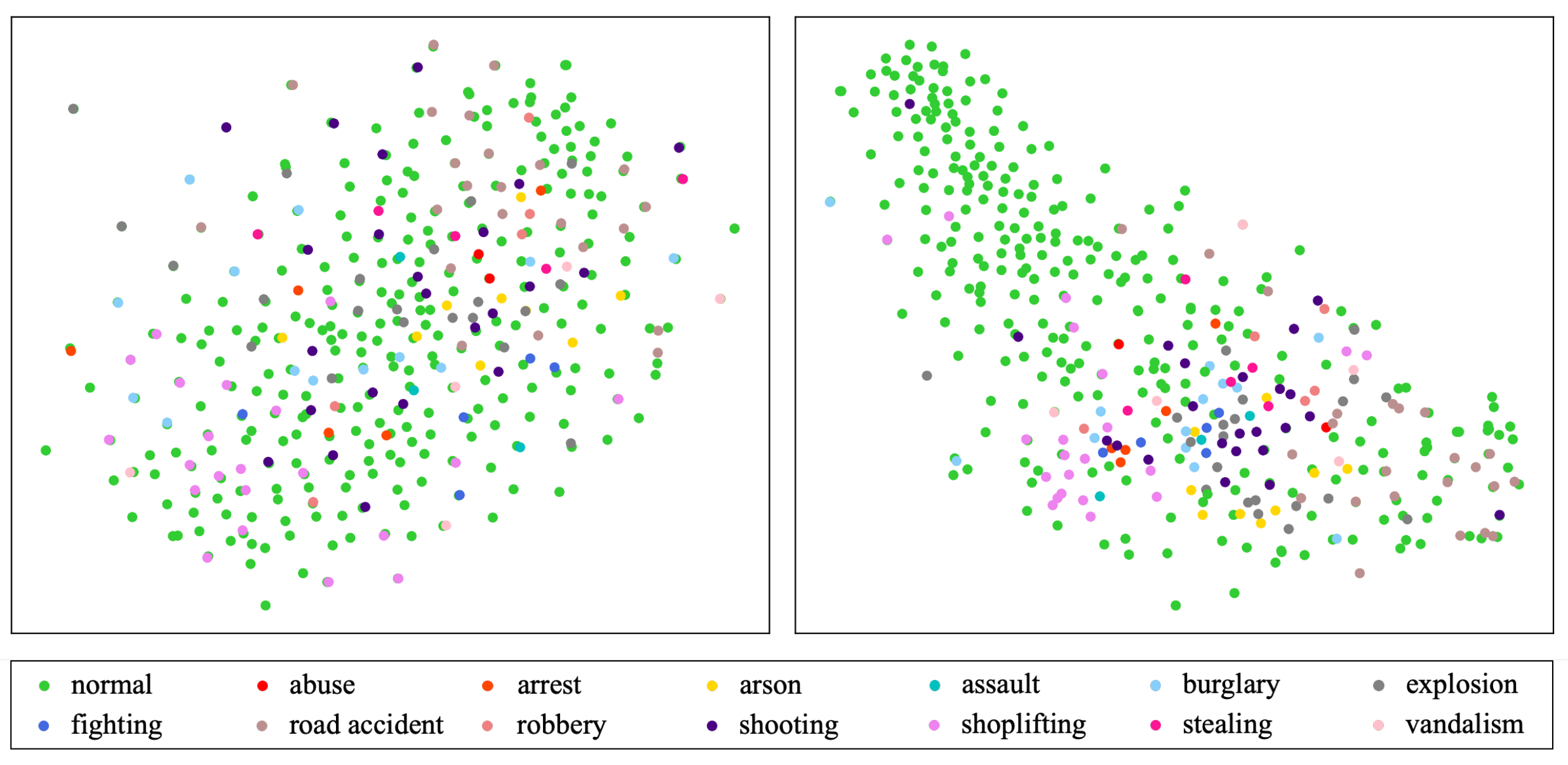}}
\hfil
\subfloat[]{\includegraphics[width=0.5\linewidth]{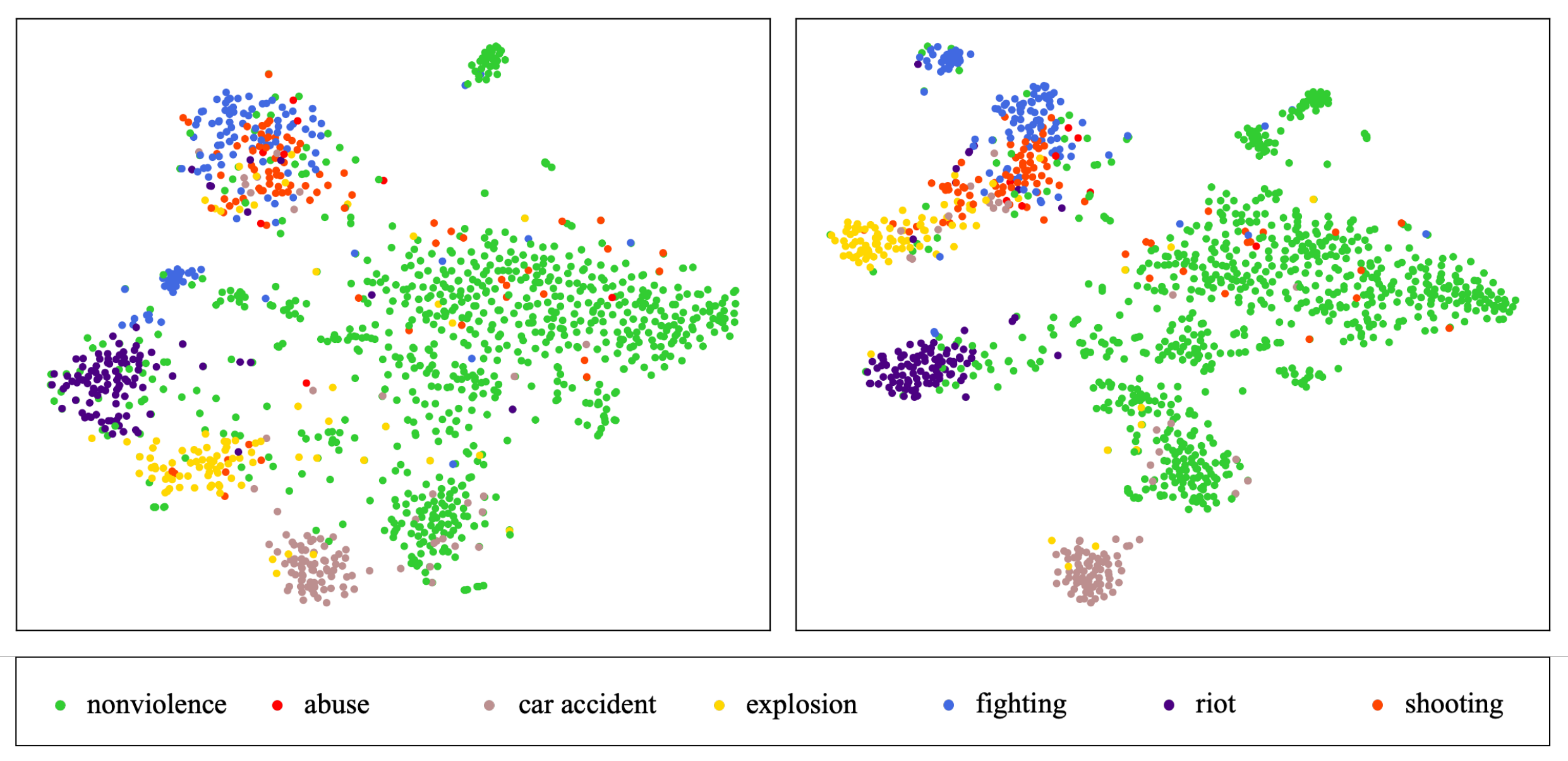}}
\caption{Distribution of discriminative features before and after PEL using t-SNE. (a) the UCF-Crime dataset. (b) the XD-Violence dataset.}
\end{figure*}

\begin{figure*}[!t]
\centering
\subfloat[Burglary024\_x264]{\includegraphics[width=0.33\linewidth]{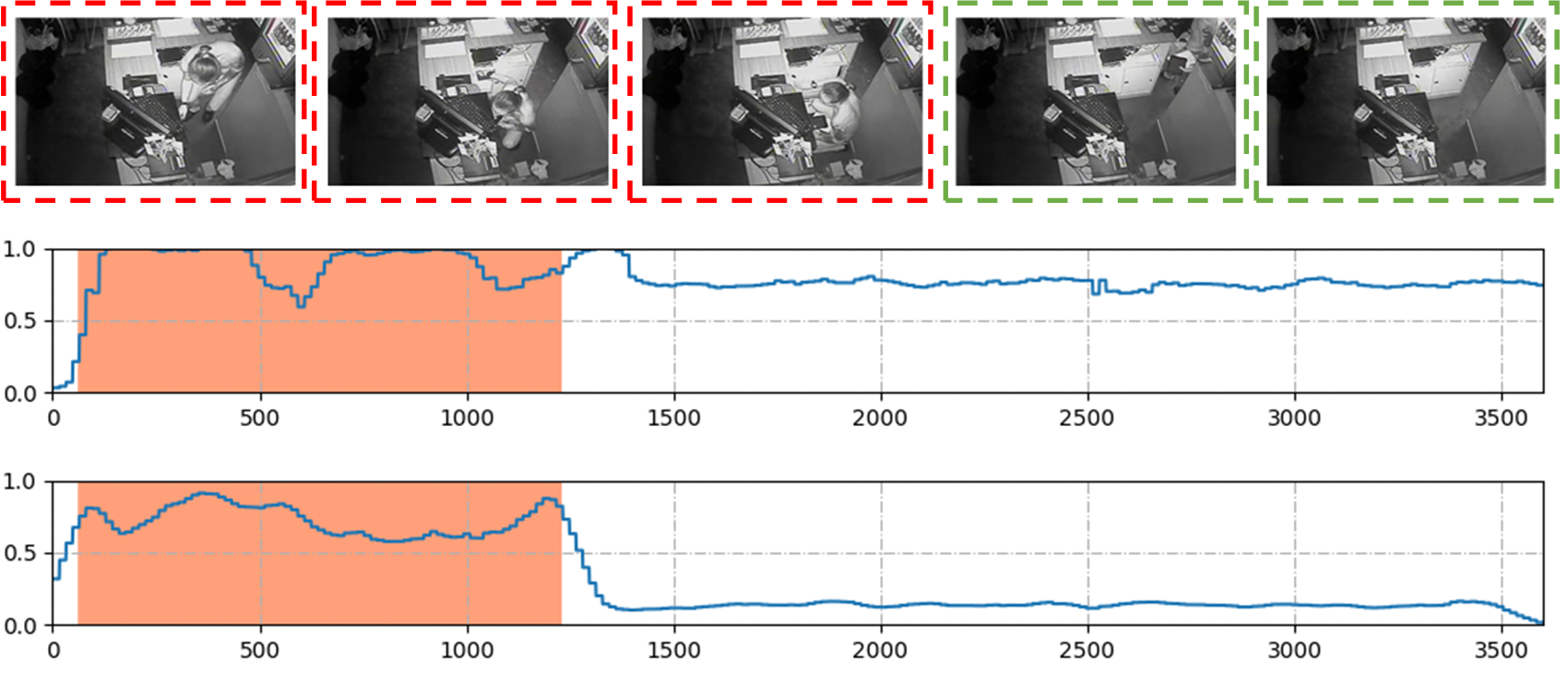}}
\hfil
\subfloat[Explosion022\_x264]{\includegraphics[width=0.33\linewidth]{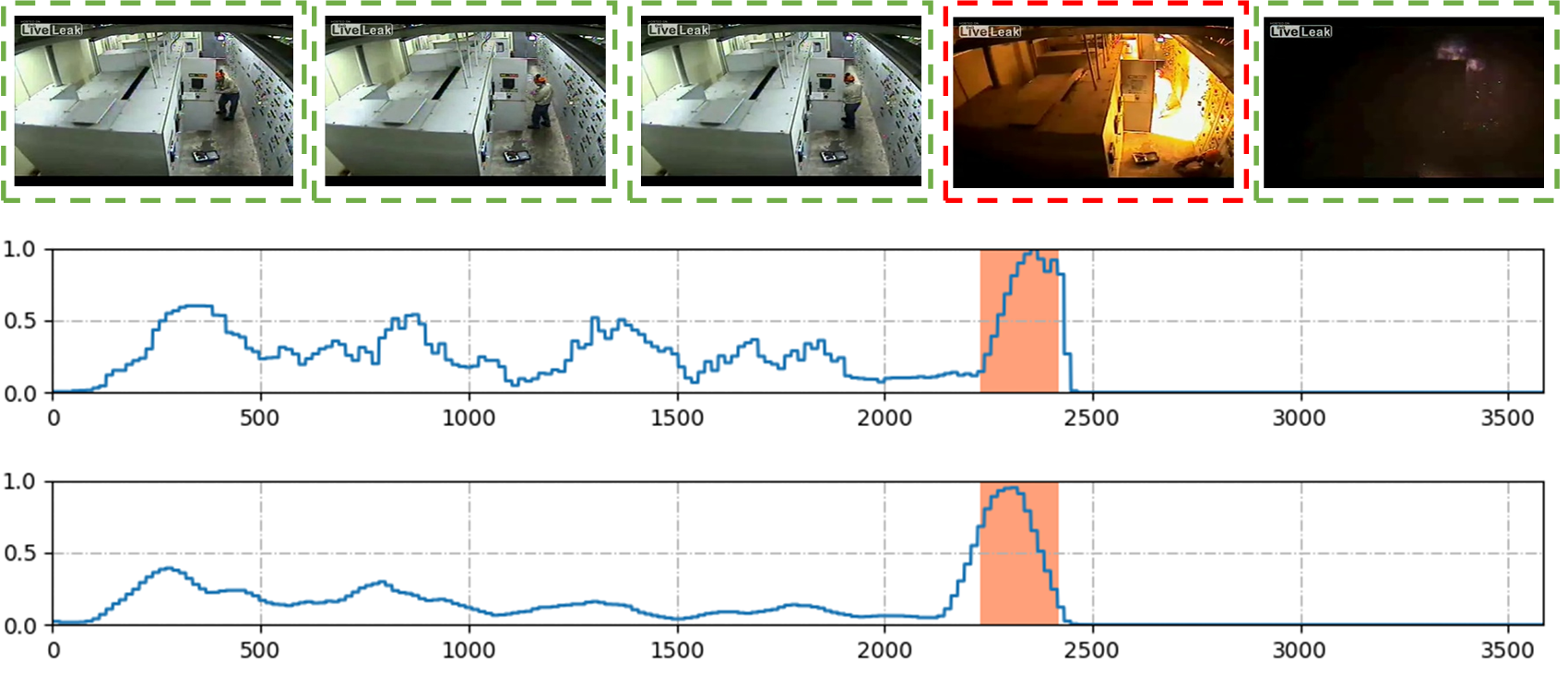}}
\hfil
\subfloat[Robbery048\_x264]{\includegraphics[width=0.33\linewidth]{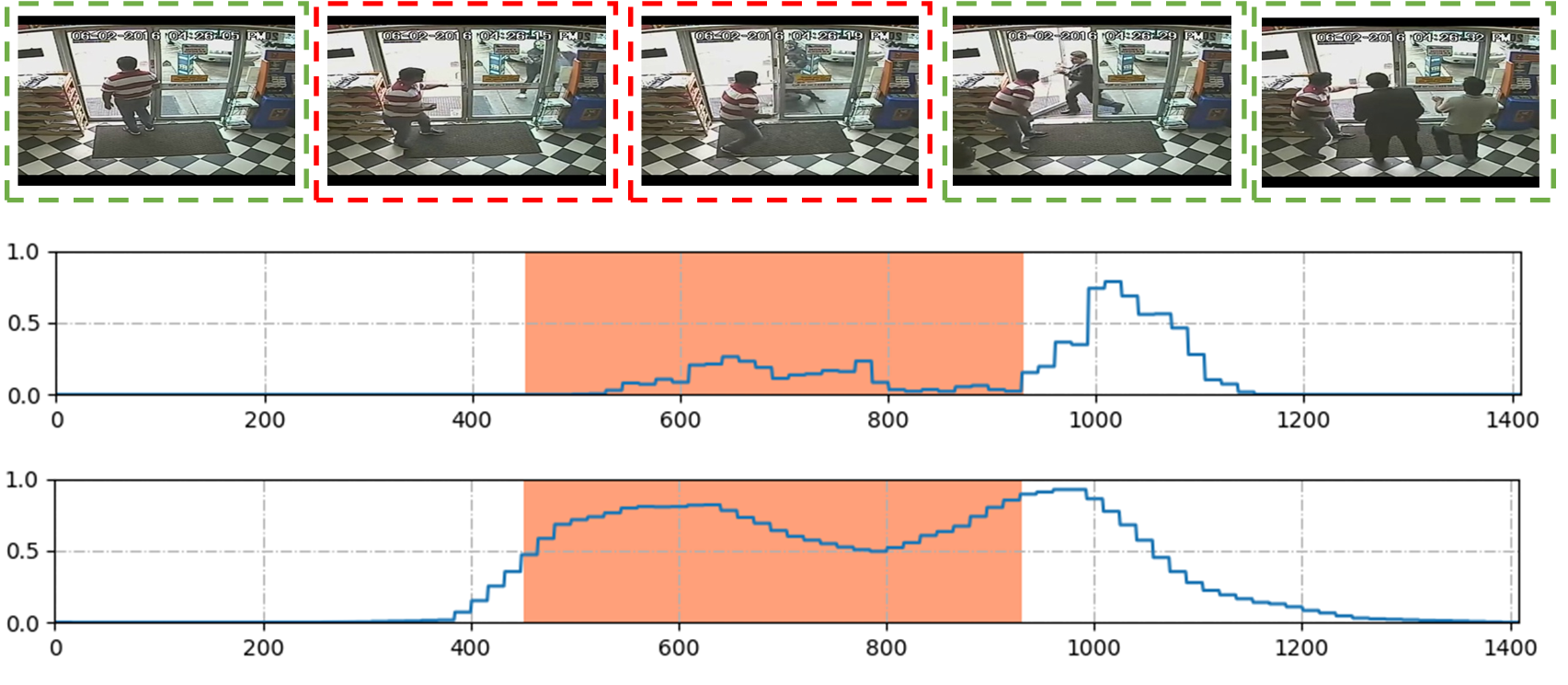}}
\hfil
\subfloat[Ip.Man.2008]{\includegraphics[width=0.33\linewidth]{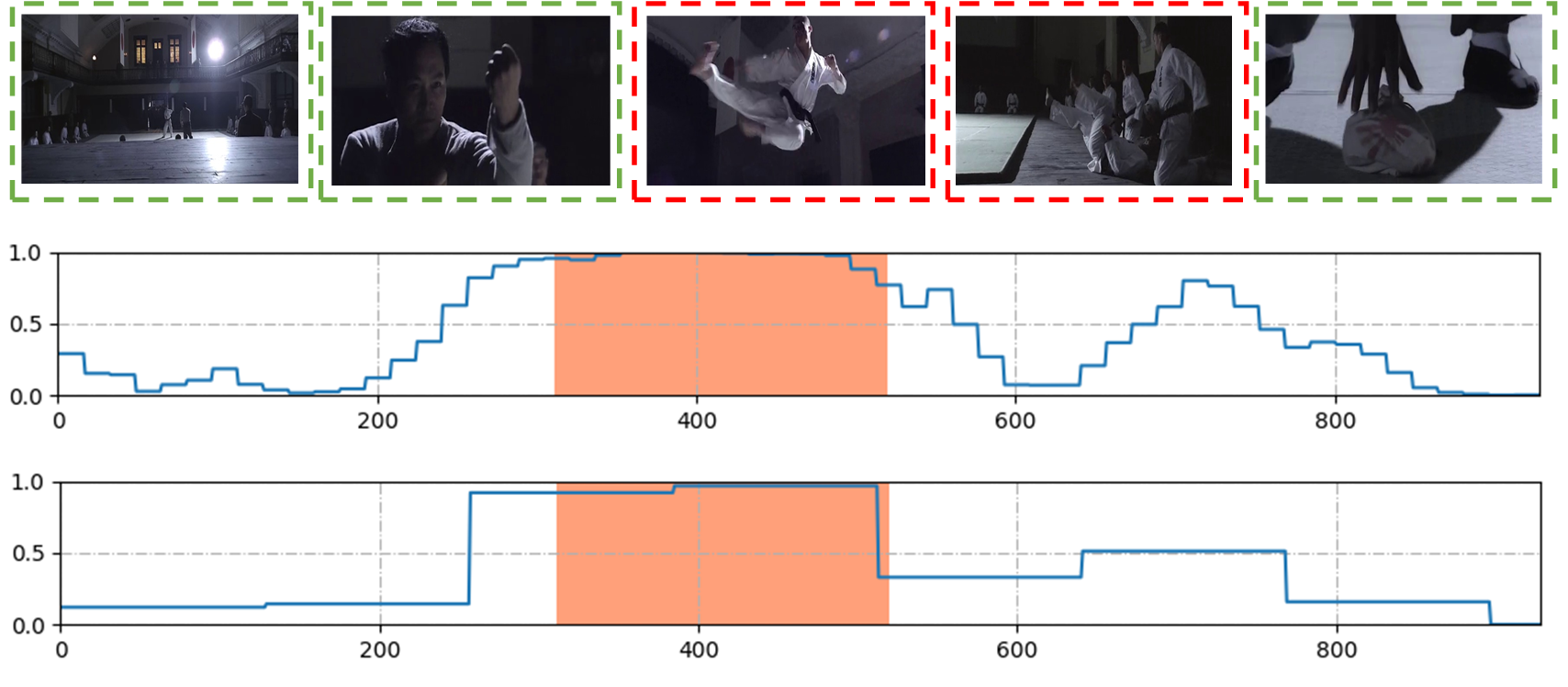}}
\hfil
\subfloat[The.Fast.and.the.Furious.2001]{\includegraphics[width=0.33\linewidth]{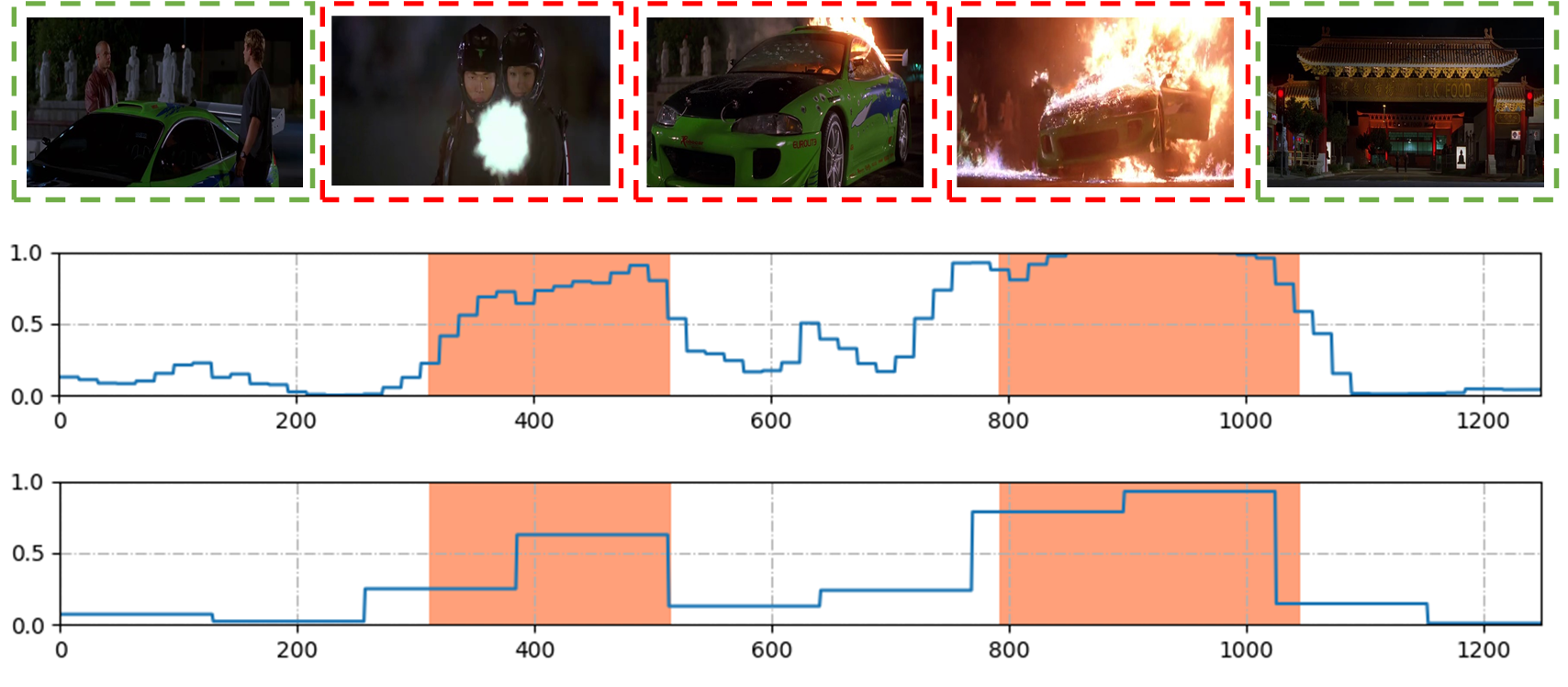}}
\hfil
\subfloat[v=0yHBkMBE8r4]{\includegraphics[width=0.33\linewidth]{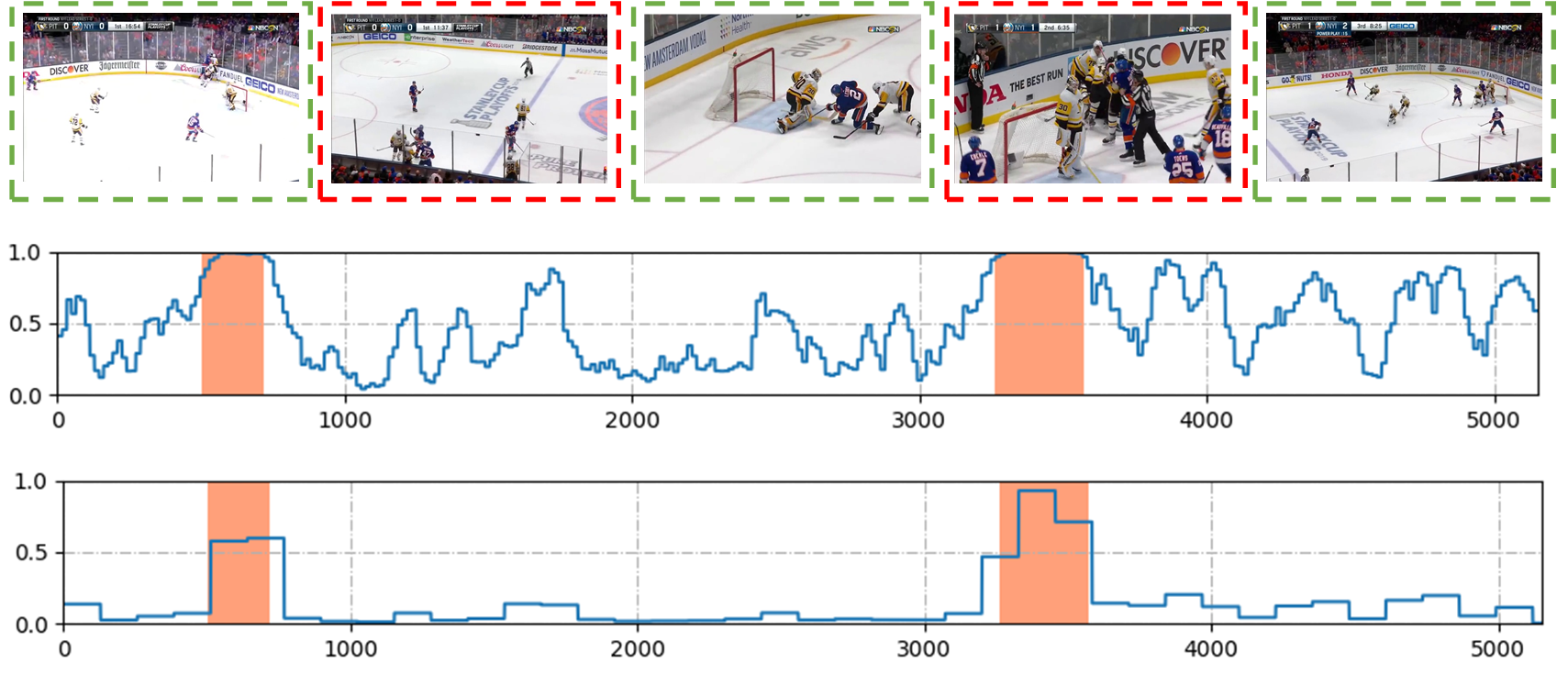}}
% \subfloat[Normal\_Videos\_783\_x264]{\includegraphics[width=0.3\linewidth]{results_fig/Normal_Videos_783_x264.png}}
% \hfil
% \subfloat[Normal\_Videos\_887\_x264]{\includegraphics[width=0.3\linewidth]{results_fig/Normal_Videos_887_x264.png}}
% \hfil
% \subfloat[Normal\_Videos\_725\_x264]{\includegraphics[width=0.3\linewidth]{results_fig/Normal_Videos_725_x264.png}}
\caption{Anomaly score visualization of the proposed method. Orange regions indicate ground-truths and blue curves indicate anomaly scores. (a)-(c) are videos from UCF-Crime and (d)-(f) are from XD-Violence, where the second and third rows indicate the results of `w/ TCA' and `w/ PEL \& SS', respectively.}
\label{fig_sim}
\end{figure*}

\begin{figure*}[!t]
\centering
\subfloat[Assault010\_x264]{\includegraphics[width=0.33\linewidth]{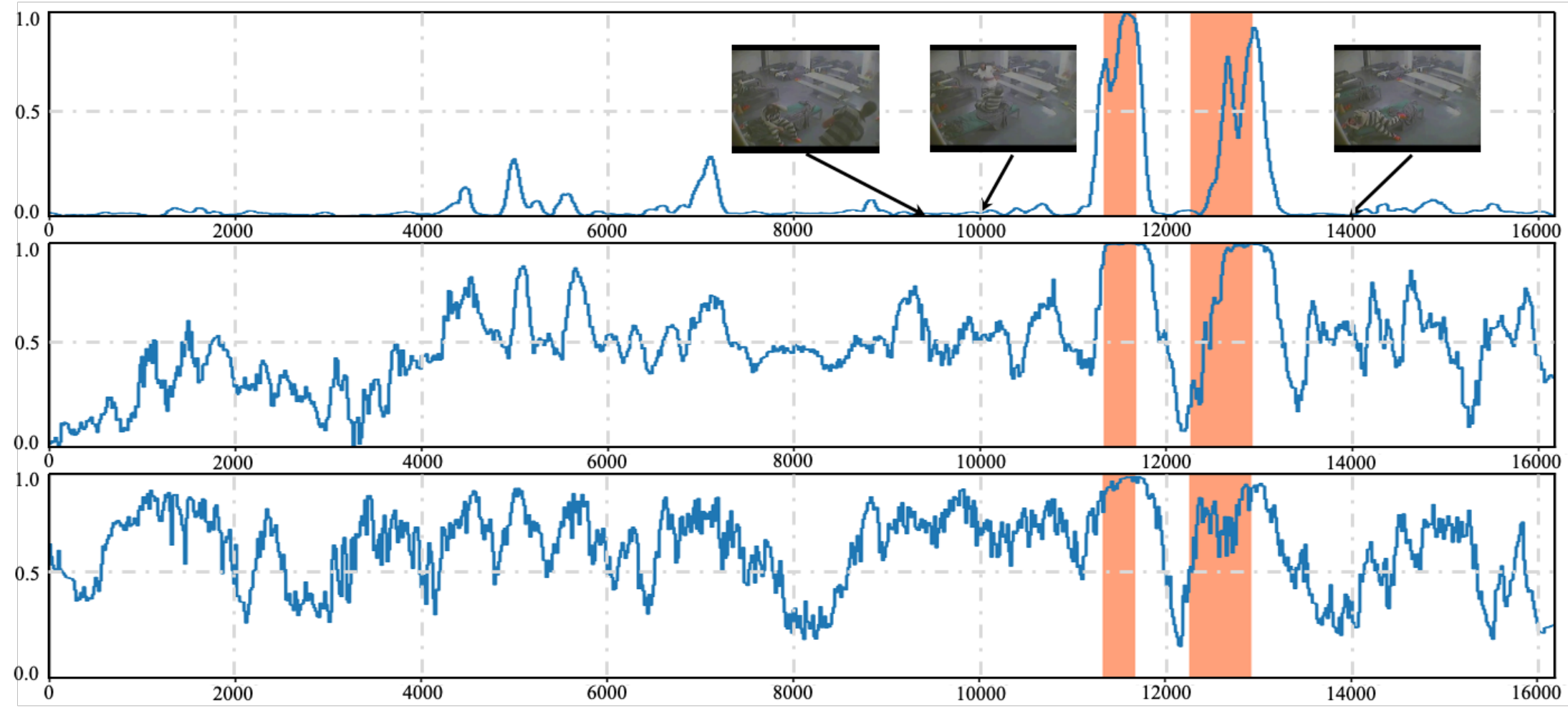}}
\hfil
\subfloat[Explosion027\_x264]{\includegraphics[width=0.33\linewidth]{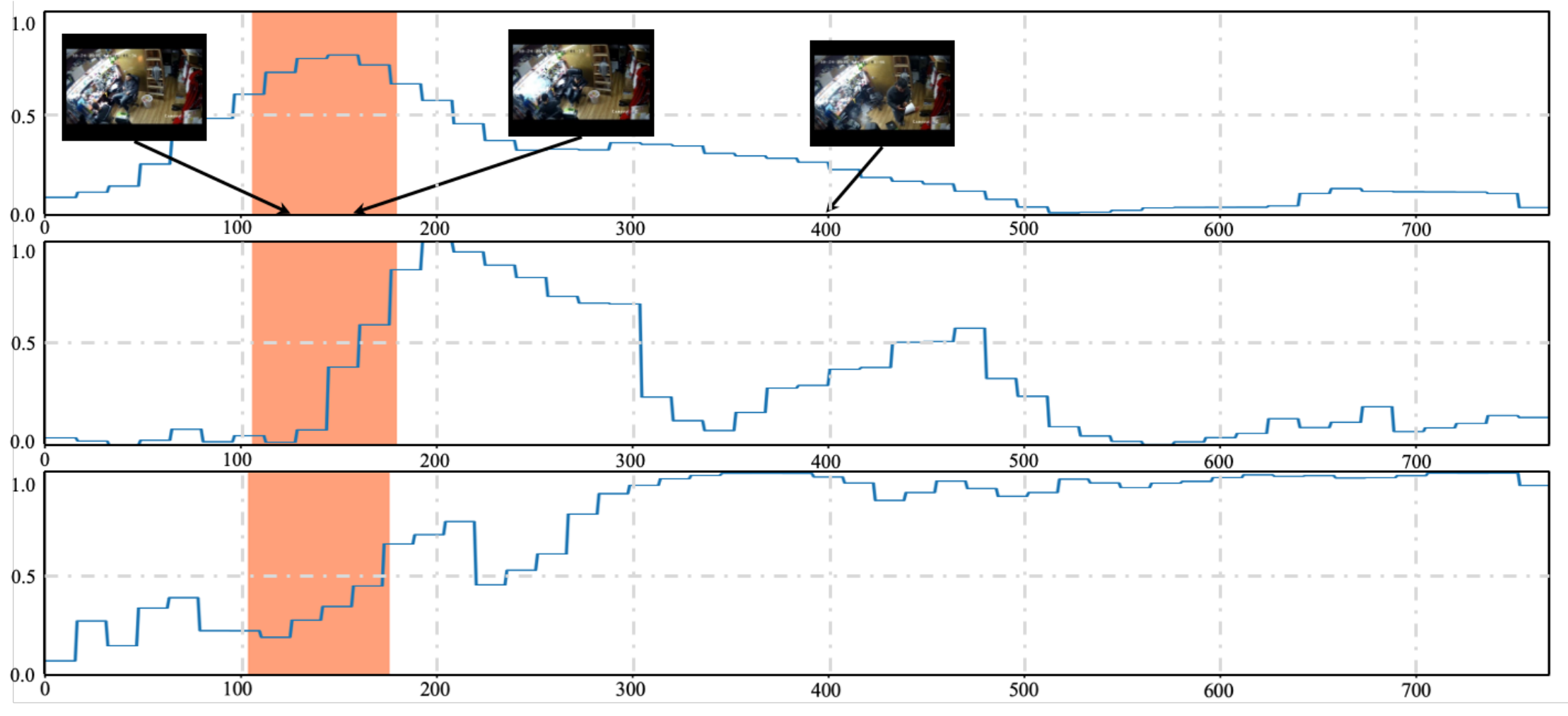}}
\hfil
\subfloat[Fighting003\_x264]{\includegraphics[width=0.33\linewidth]{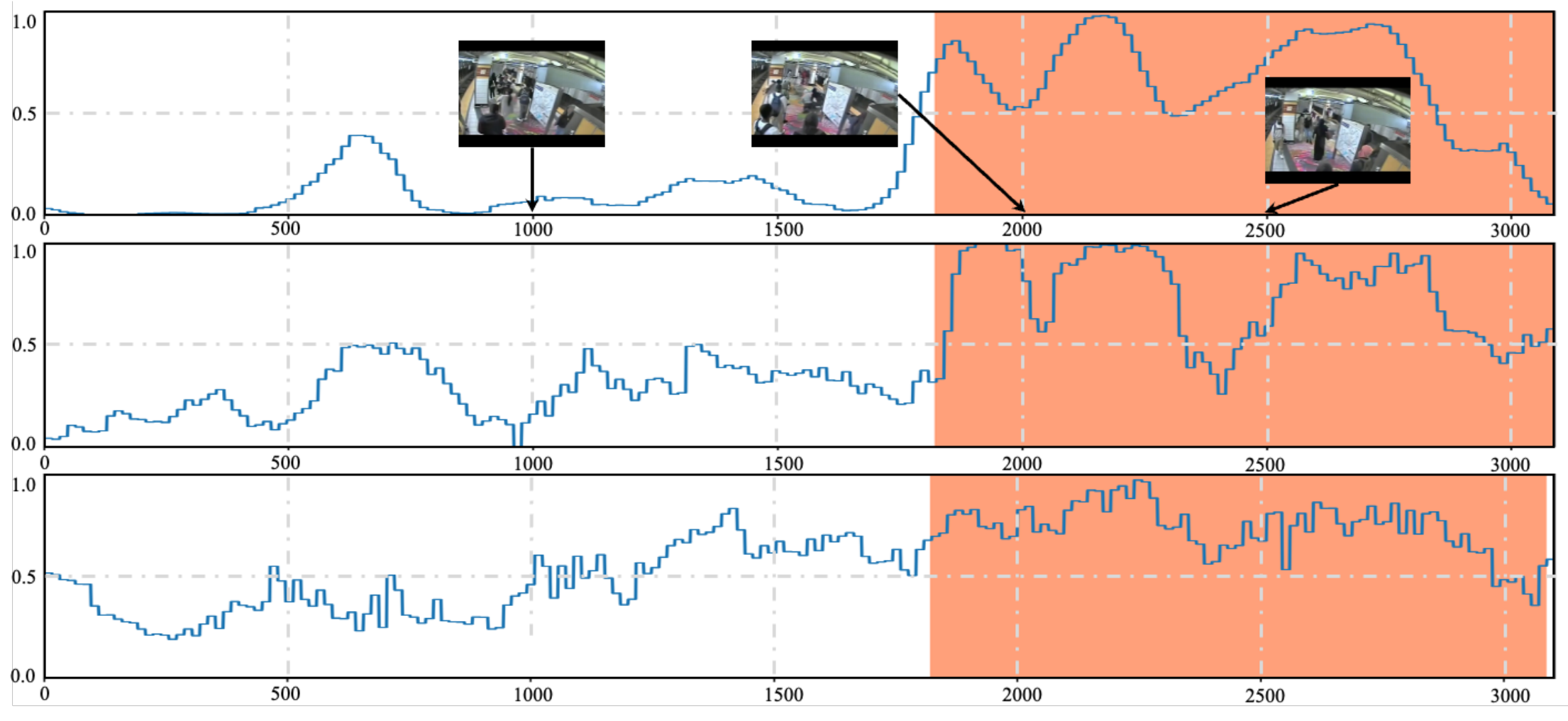}}
% \subfloat[Ip.Man.2008]{\includegraphics[width=0.3\linewidth]{results_fig/Ip.Man.2008.png}}
% \hfil
% \subfloat[The.Fast.and.the.Furious.2001]{\includegraphics[width=0.3\linewidth]{results_fig/The.Fast.and.the.Furious.2001.png}}
% \hfil
% \subfloat[v=0yHBkMBE8r4]{\includegraphics[width=0.3\linewidth]{results_fig/v=0yHBkMBE8r4.png}}
% \hfil
% \subfloat[Bullet.in.the.Head.1990]{\includegraphics[width=0.3\linewidth]{results_fig/Bullet.in.the.Head.1990.png}}
% \hfil
% \subfloat[City.Of.Men.2007]{\includegraphics[width=0.3\linewidth]{results_fig/City.Of.Men.2007.png}}
% \hfil
% \subfloat[v=lVJVRywgmYM]{\includegraphics[width=0.3\linewidth]{results_fig/v=lVJVRywgmYM.png}}
\caption{Anomaly score visualization of different methods on the UCF-Crime dataset. The first row is from our proposed framework, and the second and third rows show the detection results of  DDL\cite{Pu2022} and CUPL\cite{zhang2023exploiting}, respectively.}
\label{fig_sim}
\end{figure*}

\subsection{Parameter Evaluation}
In this subsection, we explore the effect of different hyperparameter settings on model performance.
\subsubsection{Local Window Size in TCA module}
Figure 6(a) demonstrates the effect of different local window sizes in the TCA module. The performance of the model rises and then falls on both datasets as the window continues to increase, with the best results achieved on both datasets when the window $w$ is set to 9. In general, smaller windows lack sufficient contextual information, while larger windows introduce long-range noise, leading to oscillations in false alarm rates.

\subsubsection{Kernel Size $\Delta t$ in Classifier}
Figure 6(b) illustrates the impact of varying causal convolution kernel sizes on the UCF-Crime dataset. It is observed that the best results are achieved with a kernel size of 9. However, at this kernel size, the false alarm rate is not optimal, indicating a higher number of normal samples being misclassified. This suggests a need for further refinement in the model to balance detection accuracy and false alarm rates effectively.

\subsubsection{Pooling Size $\kappa$ for Score Smoothing}
We also investigate the effect of different pooling window sizes in the SS strategy, as presented in Figure 6 (c) and (d). The results show that our model is more sensitive to the size of the sliding window compared to the moving window. The optimal performance on the UCF-Crime dataset is achieved when the sliding window size is set to 7, resulting in a false alarm rate of 0.43\%. For the XD-Violence dataset, a moving window size of 9 works best, with a further reduction in the FAR to 0.57\%.

\begin{table}[!t]
\renewcommand\arraystretch{1.2}
\setlength{\tabcolsep}{5pt}
\caption{Performance comparison of different temperature coefficients $\tau$}
\centering
\begin{tabular}{ccccccc}
\toprule[1pt]
$\tau$            & 0.01  & 0.03  & 0.05           & 0.07  & 0.09           & 0.1   \\ \hline
UCF@AUC (\%) & 85.49 & 85.94 & 85.84          & 85.61 & \textbf{86.36} & 85.98 \\
XD@AP (\%)   & 84.11 & 83.56 & \textbf{85.26} & 84.16 & 84             & 84.63 \\ \bottomrule[1pt]
\end{tabular}
\end{table}

\begin{table}[!t]
\renewcommand\arraystretch{1.2}
\caption{Performance comparison of different loss weight $\lambda$}
\centering
\begin{tabular}{ccccccc}
\toprule[1pt]
$\lambda$            & 0.01  & 0.1   & 0.5   & 1              & 5     & 10    \\ \hline
UCF@AUC (\%) & 85.11 & 85.19 & 85.52 & \textbf{86.36} & 84.51 & 84.42 \\
XD@AP (\%)   & 82.77 & 83.3  & 84.48 & \textbf{85.26} & 82.65 & 83.43 \\ \bottomrule[1pt]
\end{tabular}
\end{table}

\subsubsection{Temperature Coefficient $\tau$ and Loss Weight $\lambda$}
Tables \uppercase\expandafter{\romannumeral13} and \uppercase\expandafter{\romannumeral14} report the model's performance for different loss weights and temperature coefficients. Our model achieves optimal performance on both datasets when the weight is set to 1, which ensures a good balance of classification and alignment terms. However, the temperature coefficient varies for each dataset, with the best performance on UCF-Crime achieved at 0.09 and on XD-Violence at 0.05.

% These results suggest that the model has different levels of confidence in the positive pairs in the two datasets.

\subsection{Qualitative Analysis}
 We first use t-SNE visualization for the MLP's middle layer features, as shown in Figure 7. Before introducing PEL, the UCF-Crime dataset samples appear disordered, while the XD-Violence dataset shows clearer clustering. This difference is attributed to the temporal coherence in surveillance videos and distinct shot boundaries in XD-Violence content. Cross-modal alignment loss makes abnormal foregrounds converge towards prompt features, forming tighter clusters. It also aligns hard-negative snippets and backgrounds with a normal center, increasing the distance between abnormal and normal snippets in the embedding space. This results in a more effective differentiation of fine-grained anomalous samples compared to traditional binary classification models.

Figure 8 illustrates the anomaly scores obtained using our methods. The TCA module's detection results are shown in the second row, while the combined effect of PEL and SS is presented in the third row. PEL and SS, when used together, achieve more precise anomaly localization than TCA alone. This increased accuracy is attributed to the effective suppression of non-anomaly noise by PEL's context separation and the mitigation of false alarms caused by frame changes and luminance variations. Moreover, the integration of rich semantic information from prompts significantly enhances the model's versatility in detecting various anomalous events across different scenarios.

Finally, we compare the detection effectiveness of various methods in Figure 9. The PEL method exhibits superior sensitivity and specificity in identifying anomalies within the UCF-Crime dataset, particularly evident in scenarios like Assault, Explosion, and Fighting. PEL's anomaly scoring is characterized by sharp peaks and clear distinctions of anomalous events, in stark contrast to the broader, less defined detection of DDL and the delayed or dispersed responses of CUPL. This demonstrates PEL's exceptional temporal accuracy and event localization capabilities, establishing it as a highly effective tool for real-world surveillance applications.

% \subsection{Citations to the Bibliography}
% The coding for the citations is made with the \LaTeX\ $\backslash${\tt{cite}} command. 
% This will display as: see \cite{ref1}.

% For multiple citations code as follows: {\tt{$\backslash$cite\{ref1,ref2,ref3\}}}
%  which will produce \cite{ref1,ref2,ref3}. For reference ranges that are not consecutive code as {\tt{$\backslash$cite\{ref1,ref2,ref3,ref9\}}} which will produce  \cite{ref1,ref2,ref3,ref9}

\section{Conclusion}
In this paper, we focus on efficient temporal context modeling and semantic enhancement of visual features for weakly supervised video anomaly detection. 
We introduce a temporal context aggregation module that reuses the similarity matrix to capture local-global dependencies simultaneously. This method not only cuts down on parameters and computational load but also boosts detection capabilities, marking a significant shift from conventional paralleled systems and enhancing real-world applicability and efficiency. Recognizing the complex nature of anomalies, our model doesn’t just detect but understand  them by integrating external knowledge for semantic interpretability. This aligns the model with human cognition, offering nuanced understanding be yond typic al binary constraints. Our proposed prompt-enhanced learning module further refines this by using class-specific prompts derived from external knowledge, improving the distinction between abnormal sub-classes and maintaining clear inter-class separation. In the future, multimodal information such as motion and audio remains to be explored, and open-set anomaly detection also deserves attention.

\bibliographystyle{IEEEtran}
\bibliography{reference}

\newpage

 

% \end{IEEEbiography}



% \vfill

\end{document}